\documentclass[a4paper]{report}
\usepackage{setspace}

\usepackage{parskip}
\usepackage{amssymb,graphicx,color}
\usepackage{amsfonts}
\usepackage{latexsym}
\usepackage{amsmath}
\usepackage[a4paper, margin = 3cm, bottom = 2.5cm]{geometry}

\usepackage{url}
\usepackage{natbib}

\title{
    {\vspace{-14em} \includegraphics[scale=0.4]{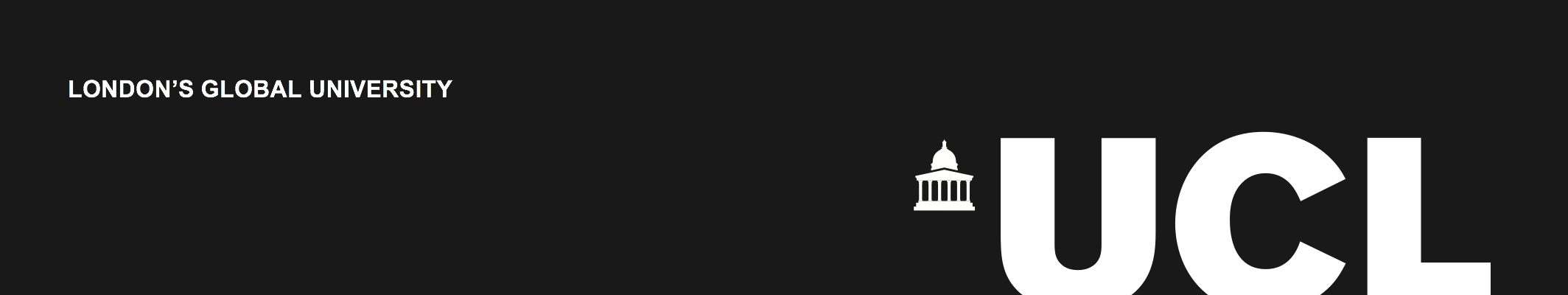}}\\
    {{\Huge Class-wise Activation Unravelling the Engima of Deep Double Descent}}\\
    {\large In-depth Analysis from an Empirical Perspective}\\
}

\date{Submission Date: April 26th, 2024}

\author{
        Candidate Number: Yufei Gu
        \thanks{
        {\textbf{Disclaimer:} This report is submitted as part of the requirement for the M.Eng. Computer Science program at UCL. It is substantially the result of my work except where explicitly indicated in the text. The report will be distributed to the internal and external examiners, but thereafter \textbf{may not} be copied or distributed except with permission from the author.}
    } \\ \\
    M.E.ng. Computer Science\\ \\
    Internal Supervisor: Tomaso Aste\\
    External Supervisor: Xiaoqing Zheng\thanks{
        The external supervisor has guided and cooperated with the author on Chapter 4 Interpolation of Noise Data in the Feature Space as detailed in \cite{gu2023unraveling}.
    }\\ \\
}

\begin{document}
 
\onehalfspacing
\maketitle

\begin{abstract}
    Double descent presents a counter-intuitive aspect within the machine learning domain, and researchers have observed its manifestation in various models and tasks. While some theoretical explanations have been proposed for this phenomenon in specific contexts, an accepted theory for its occurring mechanism in deep learning remains yet to be established. In this study, we revisited the phenomenon of double descent and discussed the conditions of its occurrence. This paper introduces the concept of class-activation matrices and a methodology for estimating the effective complexity of functions, on which we unveil that over-parameterized models exhibit more distinct and simpler class patterns in hidden activations compared to under-parameterized ones. We further looked into the interpolation of noisy labelled data among clean representations and demonstrated overfitting w.r.t. expressive capacity. By comprehensively analysing hypotheses and presenting corresponding empirical evidence that either validates or contradicts these hypotheses, we aim to provide fresh insights into the phenomenon of double descent and benign over-parameterization and facilitate future explorations. By comprehensively studying different hypotheses and the corresponding empirical evidence either supports or challenges these hypotheses, our goal is to offer new insights into the phenomena of double descent and benign over-parameterization, thereby enabling further explorations in the field.
    The source code is available at \url{https://github.com/Yufei-Gu-451/sparse-generalization.git}.
    
\end{abstract}

\setcounter{page}{1}
\tableofcontents

\chapter{Introduction}
As a contradiction to the traditional statistics' intuition, \textbf{`double descent'} has been an intriguing phenomenon in recent ages in the field of machine learning. It mainly describes the phenomenon that the generalization error of one machine learning model undergoes a trend of first decrease, then increase, and \textbf{decrease again} w.r.t. increase in model capacity (with no explicit regularization techniques!). Explorations and research from both theoretical and empirical perspectives have been proposed to bridge the gap between the `old' understanding and modern practices that have consistently demonstrated promise, particularly in the context of super-large models. However, the underlying mechanism of this phenomenon persists as a question with no answer being widely accepted. In this paper, we will re-examine this phenomenon in the scope of deep learning from an empirical perspective, by analyzing the intrinsic dynamics of various neural architectures of different scopes and attempting to provide novel insights into this question.

\section{Background}

As a well-known property of machine learning, the bias-variance trade-off suggested that the variance of the parameter estimated across samples can be reduced by increasing the bias in the estimated parameters~\citep{geman1992neural}. When a model is under-parameterized, there exists a combination of high bias and low variance. Under this circumstance, the model becomes trapped in under-fitting, signifying its inability to grasp the fundamental underlying structures present in the data. When a model is over-parameterized, there exists a combination of low bias and high variance. This is usually due to the model over-fitting noisy and unrepresentative data in the training set. Traditional statistical intuition suggests a difficulty in reducing bias and variance simultaneously and advises finding a `sweet spot' between underfitting and overfitting. This conjugate property prevents machine learning estimators from generalizing well beyond their dataset~\citep{hastie2009elements}. However, modern machine learning practices have achieved remarkable success with the training of highly parameterized sophisticated models that can precisely fit the training data (`interpolate' the training set) while attaining impressive performance on unseen test data.

To reconcile this contradiction between the classical bias-variance trade-off theory and the modern practice, \citet{belkin2019reconciling} presented a unified generalization performance curve called `Double Descent'. The novel theory posits that, with an increase in model width, the test error initially follows the conventional `U-shaped curve', reaching an interpolation threshold where the model precisely fits the training data; Beyond this point, the error begins to decrease, challenging the traditional understanding and presenting a new perspective on model behavior. Similar behaviour was previously observed in \citet{vallet1989linear,opper1995statistical,opper2001learning}, but it had been largely overlooked until the double descent concept was formally raised. 

The double descent phenomenon is prevalent across various machine learning models. In the context of linear models, a series of empirical and theoretical studies have emerged to analyze this phenomenon~\citep{bartlett2020benign, loog2020brief, belkin2020two, deng2022model, hastie2022surprises}. In deep learning, many researchers have offered additional evidence supporting the presence and prevalence of the double descent phenomenon across a broader range of deep learning models, including CNNs, ResNets, and Transformers, in diverse experimental setups~\citep{nakkiran2021deep, chang2021provable, liu2022robust, gamba2022deep}. \citet{nakkiran2021deep} specifically concluded that this phenomenon is most pronounced when dealing with label noise in the training set, while can manifest even without label noise for the first time. Despite the widely accepted empirical observation of the double descent phenomenon within the domain of deep learning, comprehending the underlying mechanism of this occurrence in deep neural networks continues to pose a significant question. 

It is well-known that deep neural networks accomplish end-to-end learning frameworks through integrating feature learning with classifier training~\citep{zhou2021over}. A neural network tailored for classification tasks can be divided into two components. The initial segment is dedicated to a feature space transformation process by selecting information pertinent to the output while discarding information associated with the input. Specifically, the conversion of the original feature space represented by the input layer to the final feature space represented by the final representation layer is referred to as the learned feature space, which is shaped by the acquired representations within fully-trained neural networks. The subsequent stage typically entails constructing a classifier based on this learned feature space, intending to complete machine-learning tasks.

The initial proposal of the double descent phenomenon as a general concept was made by \citet{belkin2019reconciling} for increasing model size (model-wise). However, \citet{nakkiran2021deep} have also shown a similar trend w.r.t. size of the training dataset (sample-wise) and training time (epoch-wise). The research further reports observation of all forms of double descent most strongly in settings with label noise in the train set. More recently, \citet{chaudhary2023parameter} revisited the phenomenon and proposed that double descent may not inherently characterize deep learning and \citet{xue2022investigating} further argues the existence of a final ascent with high label noise. While the double descent phenomenon is widely discussed in these alternative perspectives, we focus our attention on the original model-wise phenomenon in this work, briefed as `double descent' in the subsequent discussion. 

Double Descent has been shown as a robust phenomenon that occurs over diverse tasks, architectures, and optimization techniques. Subsequently, a substantial volume of research has been conducted to investigate the double descent phenomenon regarding its mechanism. A stream of theoretical literature has emerged, delving into the establishment of generalization bounds and conducting asymptotic analyses, particularly focused on linear functions~\citep{advani2020high, belkin2020two, bartlett2020benign, muthukumar2021classification, mei2022generalization, hastie2022surprises}. \citet{hastie2022surprises} highlight that when the linear regression model is misspecified, the best generalization performance can occur in the over-parameterized regime. \citet{mei2022generalization} further concluded that the double descent phenomenon can be attributed to model misspecification, which arises when there is a mismatch between the model structure and the model family. While people believe the same intuition extends to deep learning as well, nevertheless, the underlying mechanism still eludes a comprehensive understanding of this `deep' double descent phenomenon. The following section will present the related works that delve into analyzing deep double descent and their perspectives to explain the mechanism underlying this phenomenon.

\section{Related Works}

With the rise in computational power available to humans, an emerging theory of over-parameterized machine learning (TOPML) is growing, seeking to elucidate the second `descent' in the double descent phenomenon~\citep{dar2021farewell}. Research emerged on the convergence theory for over-parameterized deep learning models \citep{allen2019convergence, zou2019improved} in the first place, while we in this paper focused predominantly on the generalization aspect within TOPML. In certain research perspectives, benign over-parameterization denotes the phenomenon where larger models consistently outperform smaller ones in generalization, regardless of whether a test error peak forms at the interpolation threshold or not, as described in double descent. The inquiry posed by \citet{zhou2021over} revolves around the puzzling aspect of why over-parameterized models do not succumb to overfitting and focus on the feature space transformation aspect. 

As concluded by \citet{oneto2023we}, major theoretical perspectives within statistical learning theory encompass Complexity-based theory, Algorithmic Stability, PAC-Bayes, Differential Privacy, Compression, and Information Theory. However, not all of these perspectives apply to the benign over-parameterization phenomenon of deep learning models, and many of them still lack comprehensive exploration (given the recent discovery of this phenomenon). Most recent theoretical investigations into the problem have been conducted by deriving generalization bounds across various architectures and assumptions. These include studies on ERM and SVM classifiers \citep{wang2021benign}, ReLU networks dealing with adversarial label noise \citep{frei2021provable}, two-layer CNNs \citep{cao2022benign, kou2023benign}, and ReLU networks under lazy training \citep{zhu2023benign}. From a more intuitive standpoint, \citet{teague2022geometric} drew a connection between generalization capacity and the geometric properties of volumes within hyperspace, while \citet{poggio2022compositional} endeavoured to construct a framework on compositional sparsity, thought to enhance generalization ability. 

In addition to pure theoretical investigations, several studies with a stronger empirical focus delve into double descent or benign over-parameterization in deep learning from various perspectives: bias-variance decomposition~\citep{yang2020rethinking, chen2023bias}; samples to parameters ratio~\citep{belkin2020two, nakkiran2021deep}; decision boundaries~\citep{somepalli2022can}; and the sharpness of interpolation~\citep{gamba2022deep}. While offering an extensive array of experimental findings that replicate the double descent phenomenon, \citet{somepalli2022can} further studied the decision boundaries of random data points and established a relationship between the fragmentation of class regions and the double descent phenomenon. 

In contrast to existing studies, our approach distinguishes itself by exploring the phenomenon through the empirical analysis centring class-wise activations in neural networks trained for image classifications. Activations in neural network hidden layers can be denoted as hidden representations or hidden features of input data. The information theoretic interpretation suggested that these activations through layers are a maximally compressed mapping of the input variable that preserves as much as possible the information on the output variable \citep{tishby2015deep}. We believe there have been fewer efforts to demonstrate a connection between class-wise activations and the double descent or benign over-parameterization phenomenon. This is the focus of our exploration in this paper. Furthermore, we have also explored how corrupted labels in training data will impact the performance of neural networks' generalization ability in Chapter \ref{chapter5:interpolation}. One of the related works on this question consisted of the concurrent ones by \citet{gamba2022deep}, which measure sharpness at clean and noisily-labelled data points and empirically showed that smooth interpolation emerging both for large over-parameterized networks and trained large models confidently predicts the noisy training targets over large volumes around each training point. While their study also entailed an analysis involving separate evaluations of clean and noisily labelled data points, we employed the $k$-nearest neighbour algorithm to infer the inter-relationship between these two categories of class-wise activations of training data. In conclusion, by analyzing the properties of class-wise activations in neural architectures of different scales, we attempted to build connections with double descent and provide some fresh understanding of its forming mechanisms.

\section{Summary of Contributions}

The main contributions of this paper are summarized as follows: 
\begin{itemize}
    \item Re-examined the double descent phenomenon and discussed the conditions for its occurrence.
    \item Explored and evaluated the inherent learned class patterns within neural network activations, showcasing the complete dynamics of pattern similarities across both representation layers and classifier alterations concerning model width.
    \item Proposed an empirical methodology for estimating intrinsic model complexity by measuring class-wise activation richness which demonstrated a robust relationship to the double descent phenomenon.
    \item Presented an in-depth analysis of the interpolation strategy of noisy samples in the hidden feature space employed by various networks, explaining the overfitting phenomenon across neural architectures of different complexity.
\end{itemize}

\vspace{10pt}
The structure of this paper is outlined below: In Chapter \ref{chapter2:experiments}, we will present our study on the model-wise generalization performance of three neural architectures: fully connected neural networks (FCNNs), convolutional neural networks (CNNs), and residual neural networks (ResNets) with no explicit regularization techniques and optional external label noise. By examining these architectures under different conditions, we wish to provide a validation of the robustness of the double decent phenomenon. Furthermore, we will look into several hypotheses constructed on previous research, including perspectives from intrinsic class patterns in Chapter \ref{chapter3:activation}, associated model complexity/richness in Chapter \ref{chapter4:complexity} and the interplay between clean and noisy representations in Chapter \ref{chapter5:interpolation}. In each of these three chapters, we present certain empirical methodologies to validate the proposed hypothesis and provide certain insights into these perspectives. 
In particular, \textbf{the experimental findings discussed in Chapters \ref{chapter2:experiments} and \ref{chapter5:interpolation} have been previously published by the author though the exact details are omitted in this submission to avoid beaching anonymity.} 
Lastly, in Chapter \ref{chapter6:conclusion}, we will evaluate our exploration and discuss the questions left to be answered by future research. Additional experiments, corresponding results and discussions are provided in the Appendix.

\chapter{Model-wise Generalization Performance\label{chapter2:experiments}}
This chapter mainly focused on introducing the experiment setups utilized in this study and presented the benign over-parameterization phenomenon under these setups serving as a replication of previous studies and forming the central topic for further discussions.

\section{Experiment Setup}

This section outlined the experimental setups, encompassing the specific neural architectures utilized, the datasets employed for both training and evaluation, the training methodologies applied to the architectures and datasets, and the procedures for introducing external label noise. The detailed descriptions of the experimental setup are intended to facilitate the reproducibility of this study. Further details can be referenced through the source code available in the submission.
Further details can be referenced through the source code available at \url{https://github.com/Yufei-Gu-451/sparse-generalization.git}.

\subsection{Neural Architecture}
In this study, three major neural architectures, as discussed in previous research, are examined and outlined below:

\begin{itemize}
    \item \textbf{Fully Connected Neural Networks (FCNN):} We adopt a simple two-layer fully connected neural network with varying width $k$ on the first hidden layer, for $k$ in the range of [1, 1000]. The second fully connected layer is adopted as the classifier.
    \item \textbf{Standard Convolutional Neural Networks (CNN):} We consider a family of standard convolutional neural networks formed by 4 convolutional stages of controlled base width [$k$, $2k$, $4k$, $8k$], for $k$ in the range of [1, 64], along with a fully connected layer as the classifier. The MaxPool is [2, 2, 2, 4]. For all the convolution layers, the kernel size = 3, stride = 1, and padding = 1. This architecture implementation is adopted from \citet{nakkiran2021deep}.
    \item \textbf{Residual Neural Networks (ResNet):} We parameterize a family of ResNet18s from \citet{he2016deep} using 4 ResNet blocks, each consisting of two BatchNorm-ReLU-convolution layers. We scale the layer width (number of filters) of convolutional layers as [$k$, $2k$, $4k$, $8k$] for a varied range of $k$ within [1, 64] and the strides are [1, 2, 2, 2]. The implementation is adopted from \url{https://github.com/kuangliu/pytorch-cifar.git}.
\end{itemize}

We adopt various scaling factors for each neural architecture and compare the performance of different model sizes using the universal layer width unit $k$. Although some may anticipate viewing the experimental outcomes in terms of model parameter scales, we assert that this choice does not affect our conclusions, given the monotonic relationship between these two factors. The scaling of model parameters to the width unit $k$ is detailed in Appendix \ref{appendix:scaling} as a supplement to this view.

\subsection{Datasets}

In this study, we utilize the MNIST and CIFAR-10 datasets to train and evaluate neural architectures. MNIST (Modified National Institute of Standards and Technology database) is a handwritten digits database commonly used for image processing systems and evaluation of machine learning algorithms \citep{mnist}. The MNIST database contains 60,000 training images and 10,000 testing images, although only 4,000 are utilized for training, following the same sub-sampling approach as described in the initial report of double descent \citep{belkin2019reconciling}. All 10,000 testing images are used for evaluation in this study. All images of handwritten digits in the MNIST database are presented in black and white with grayscale and adhere to a uniform format of 28x28 pixels.

CIFAR-10 (Canadian Institute For Advanced Research) is another database of images widely used for machine learning research, with 10 representing classes: airplanes, cars, birds, cats, deer, dogs, frogs, horses, ships, and trucks \citep{cifar10}. The CIFAR-10 database consists of 60,000 images, with each class comprising 6,000 images. In this study, 50,000 images are used as training images, while the remaining 10,000 are allocated as testing images. All images are coloured with 3 channels, each channel adheres to a uniform format of 32x32 pixels. Both the MNIST and the CIFAR-10 databases provide diverse and representative samples for training and testing, serving the purpose of accessing the deep learning methodologies. 

\subsection{Training Schemes}

We describe the experiment setup along with hyper-parameters we used in the training of all neural architectures below. All networks are trained using the Stochastic Gradient Descent (SGD) optimizer with zero momentum and Cross-entropy Loss for evaluation. A separate learning rate scheme is employed for different models: $lr = \frac {0.05} {\sqrt{1 + [epoch / 50]}}$ for FCNNs and $lr = \frac {0.05} {\sqrt{1 + [epoch * 10]}}$ for CNNs and ResNet18s. For every model, the learning rate starts with an initial value of 0.05 and updates every 50 epochs. A batch size of 128 is employed, and no explicit data augmentation or regularization techniques (including dropouts or pruning) are utilized in our experimental setup. Utilizing a sequence of models trained under a uniform training scheme showcases the double descent phenomenon across different models. 

\subsection{External Label Noise}

While the presence of label errors has been established in the majority of commonly used benchmarking datasets including MNIST and CIFAR-10~\cite{zhang2017method, northcutt2021pervasive}, to facilitate comparative analysis of our findings, we intentionally introduce explicit label noise. In our experimental setup, label noise with probability $p$ signifies that during training, samples are assigned a uniformly random label with probability $p$, while possessing the correct label with probability $(1-p)$ otherwise. We interpret a neural network with input variable $x \in R^d$ and learnable parameter $\theta$, incorporating all weights and biases. The training dataset is characterized as a collection of training points $(x_n, y_n)$, where $n$ ranges from 1 to $N$. A training dataset with $N$ data points comprises $m = p \times N$ noise labelled data points and $\hat{m} = (1-p) \times N$ clean data points. 

It is important to note that the label noise is sampled only once and not per epoch, and this original and random label information is stored before each experiment. Each experiment is replicated multiple times, and the final result is obtained by averaging the outcomes. Owing to the presence of inherent label errors, the probability $p$ is solely employed to signify the extent of explicit label noise and does not reflect the overall noise level present in the dataset. Our primary experimental conclusion is drawn from comparing results obtained under a dataset with a relatively noisy level.

\section{Experiment Results}

In this section, we showcase the experimental outcomes, illustrating the performance of each neural architecture trained on MNIST and CIFAR-10 across various model scales to elucidate the impact of parameterization.

\subsection{MNIST}

\begin{figure}[h]
\begin{center}
\begin{tabular}{ccc}
    \begin{minipage}[b]{0.3\textwidth}
        \includegraphics[width=\textwidth]{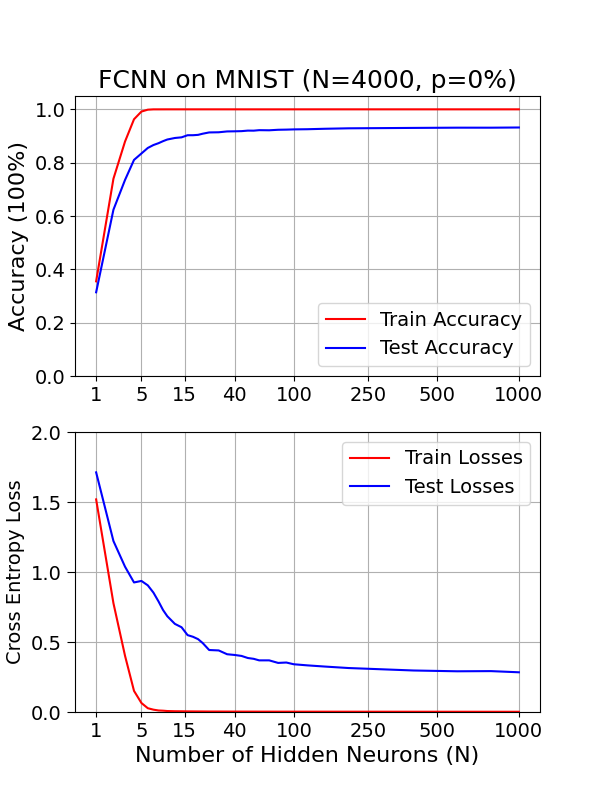}
    \end{minipage} &
    \begin{minipage}[b]{0.3\textwidth}
        \includegraphics[width=\textwidth]{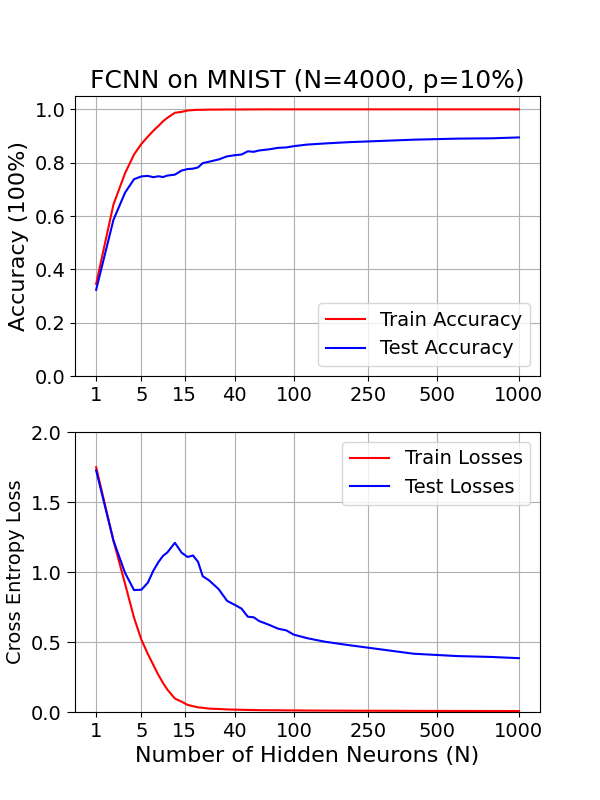}
    \end{minipage} &
    \begin{minipage}[b]{0.3\textwidth}
        \includegraphics[width=\textwidth]{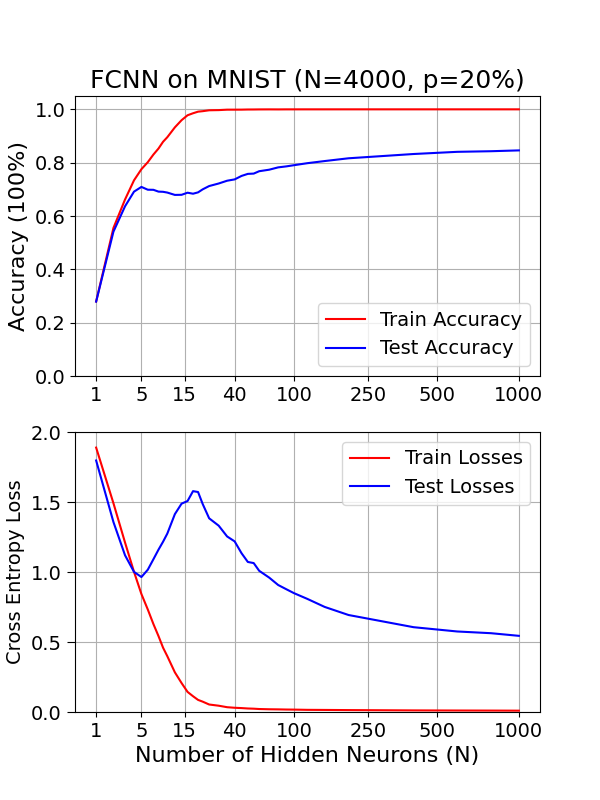}
    \end{minipage} \\
`   \fontsize{8}{8}\selectfont (a) FCNN on MNIST $(p=0\%)$ & 
    \fontsize{8}{8}\selectfont (b) FCNN on MNIST $(p=10\%)$ & 
    \fontsize{8}{8}\selectfont (c) FCNN on MNIST $(p=20\%)$ \\
\end{tabular}
\end{center}
\caption{\label{fig:MNIST-FCNN}
    The generalization performance of two-layer FCNNs trained on MNIST $(N=4000)$, under varying explicit label noise ratios of $p = [0\%, 10\%, 20\%]$. Optimized with SGD for 4000 epochs and decreasing learning rate. The test error curve with no external label noise is monotonic over the increase of model width $k$. The test error curve with external label noise performs the double descent phenomenon which decreases and increases and decreases again after the interpolation threshold.
}
\end{figure}

We commence our exploration by revisiting one of the earliest experiments that revealed the presence of the double descent phenomenon, as presented in the work of \cite{belkin2019reconciling}. This experiment involved the training of a basic two-layer FCNN on a dataset comprising 4000 samples from the MNIST dataset ($N=4000$). When zero label noise ($p=0\%$) is introduced, as depicted in Figure \ref{fig:MNIST-FCNN}(a), both the train and the test error curve decrease monotonically with increasing model width. The training error exhibits a faster decline and approaches zero at approximately $k=15$, indicating the occurrence of an interpolation threshold under these conditions. The test error still decreases after this interpolation threshold, demonstrating a benign over-parameterization phenomenon. This noticeable alteration in the shape of the test error curve at the interpolation threshold may be attributed to the intrinsic label errors of MNIST~\cite{northcutt2021confident}.

With label noise ($p=10\%/20\%$) introduced in the training dataset, a more distinguishable test error peak starts to form around the interpolation threshold as shown in Figure \ref{fig:MNIST-FCNN}(b, c). Starting with small models, the test error curve exhibits a U-shaped pattern in the first half and the bias-variance trade-off is reached at the bottom of the U-shaped curve. The decrease of both train and test error before this bias-variance trade-off indicates the model undergoes an under-fitting stage. With the expansion of the model size, the model transitions into an overfitting regime, leading to a peak in the test error around the interpolation threshold, while the train error converges to 0. Finally, after surpassing this test error peak, the testing loss experiences a decline once more, showcasing how highly over-parameterized models can ultimately yield improved generalization performance. This intriguing behaviour of a second decrease in test loss is often referred to as double descent. As the test loss undergoes a pattern of first decreasing, then increasing, and finally decreasing again, the test accuracy exhibits the opposite trend, i.e., first increasing, then decreasing, and then increasing once more. As \cite{nakkiran2021deep} reported, the test loss peak is positively correlated with the label noise ratio $p$. In other words, as the label noise ratio increases, the peak in the test loss also tends to increase accordingly.

\subsection{CIFAR-10}

\begin{figure}[h]
\begin{center}
\begin{tabular}{ccc}
    \begin{minipage}[b]{0.3\textwidth}
        \includegraphics[width=\textwidth]{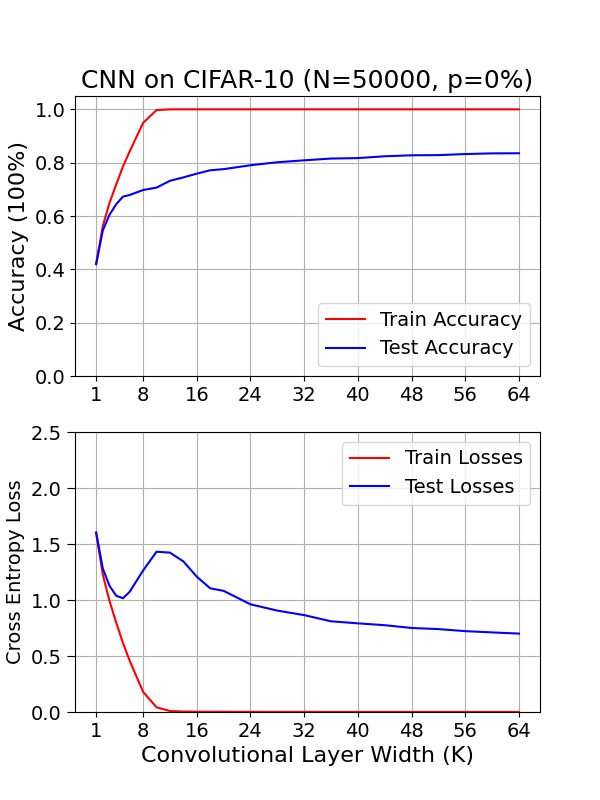}
    \end{minipage} &
    \begin{minipage}[b]{0.3\textwidth}
        \includegraphics[width=\textwidth]{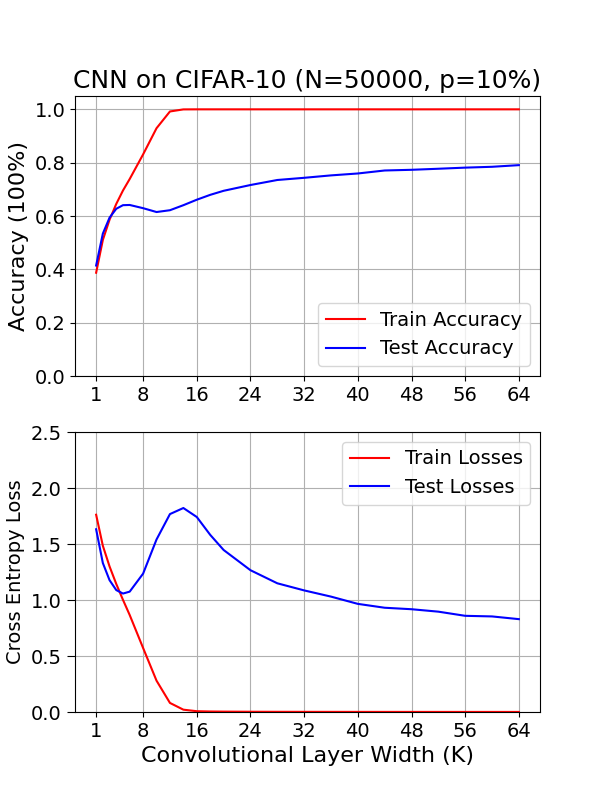}
    \end{minipage} &
    \begin{minipage}[b]{0.3\textwidth}
        \includegraphics[width=\textwidth]{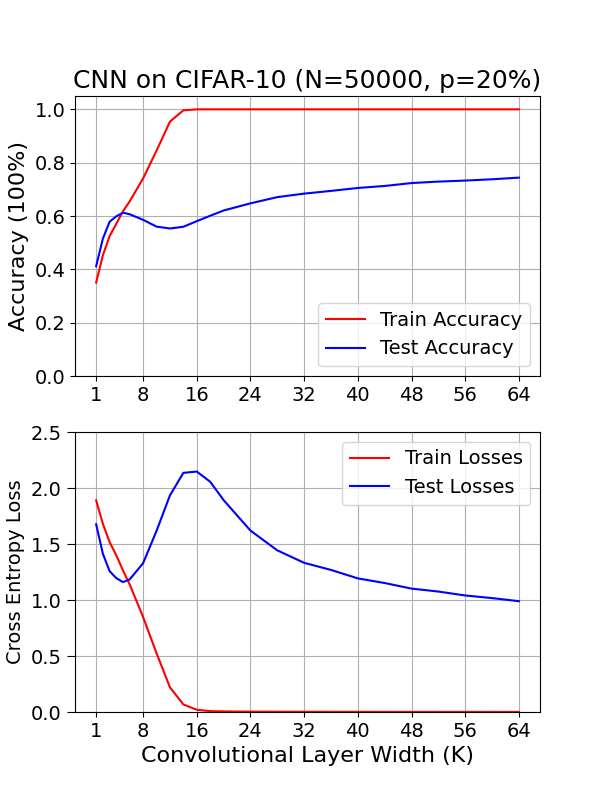}
    \end{minipage} \\
    \fontsize{8}{8}\selectfont (a) CNN on CIFAR-10 $(p=0\%)$ & 
    \fontsize{8}{8}\selectfont (b) CNN on CIFAR-10 $(p=10\%)$ & 
    \fontsize{8}{8}\selectfont (c) CNN on CIFAR-10 $(p=20\%)$ \\
\end{tabular}
\end{center}
\caption{\label{fig:CIFAR-10-CNN}
    The generalization performance of five-layer CNNs trained on CIFAR-10 ($N=50000$), under varying explicit label noise ratios of $p = [0\%, 10\%, 20\%]$. Optimized with SGD for 200 epochs and decreasing learning rate. All test error curve performs the double descent phenomenon which decreases and increases and decreases again after the interpolation threshold. 
}
\end{figure}

After a baseline case of learning with FCNNs on MNIST, we extended the experiment results to image classification on the CIFAR-10 dataset, using two famous neural network architectures: five-layer CNN and ResNet18. We first visit the experiment when CIFAR-10 ($N=50000$) with zero explicit label noise ($p=0\%$) is trained with CNNs: Figure \ref{fig:CIFAR-10-CNN}(a) demonstrated a proposed double descent peak arises around the interpolation threshold. We posit that the variance in observations could stem from a higher ratio of label errors in CIFAR-10 compared to MNIST, as suggested by research estimates~\cite{zhang2017method, northcutt2021pervasive}. Nevertheless, when additional label noise ($p=10\%/20\%$) is introduced, the test error peak still increases accordingly, proving the positive correlation between noise and double descent according to Figure \ref{fig:CIFAR-10-CNN}(b,c). While label noise increases the difficulties of models fitting the train set and shifts the interpolation threshold rightwards, the test error peak also shifts correspondingly. 

\begin{figure}[h]
\begin{center}
\begin{tabular}{ccc}
    \begin{minipage}[b]{0.3\textwidth}
        \includegraphics[width=\textwidth]{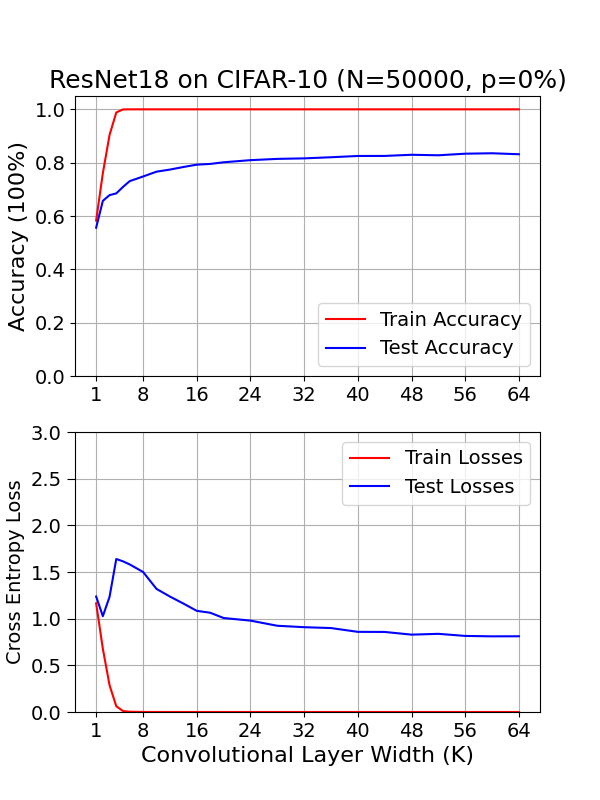}
    \end{minipage} &
    \begin{minipage}[b]{0.3\textwidth}
        \includegraphics[width=\textwidth]{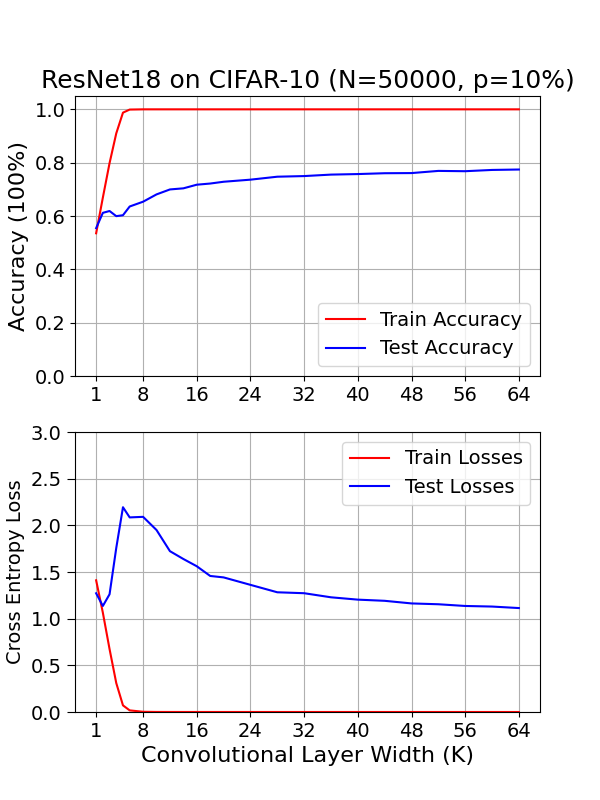}
    \end{minipage} &
    \begin{minipage}[b]{0.3\textwidth}
        \includegraphics[width=\textwidth]{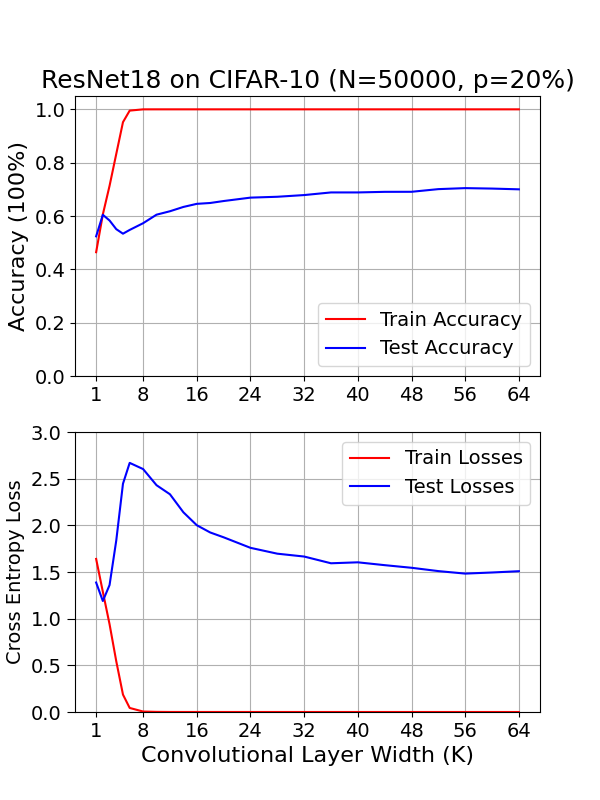}
    \end{minipage} \\
    \fontsize{8}{8}\selectfont (a) ResNet on CIFAR-10 $(p=0\%)$ & 
    \fontsize{8}{8}\selectfont (b) ResNet on CIFAR-10 $(p=10\%)$ & 
    \fontsize{8}{8}\selectfont (c) ResNet on CIFAR-10 $(p=20\%)$ \\
\end{tabular}
\end{center}
\caption{\label{fig:CIFAR-10-ResNet18}
    The generalization performance of ResNet18s trained on CIFAR-10 ($N=50000$), under varying explicit label noise ratios of $p = [0\%, 10\%, 20\%]$. Optimized with SGD for 200 epochs and decreasing learning rate. All test error curve performs the double descent phenomenon which decreases and increases and decreases again after the interpolation threshold. 
}
\end{figure}

Finally, with ResNet18 trained on CIFAR-10 ($N=50000$) in Figure \ref{fig:CIFAR-10-ResNet18}, we can see that the double descent phenomenon is robust w.r.t. model size and across different noise ratios. Compared to five-layer CNNs, the increased depth of convolutional layers and reduced training difficulty by the introduction of residual connections lead to an earlier interpolation threshold concerning the same model width unit $k$. Thus, the under-parameterized regime is compressed, and the initial decrease in test error caused by underfitting lasts shortly. However, by comparing the model's performance in the over-parameterized regime, we may observe that while ResNets outperform CNNs of intermediate scale initially, the introduction of additional label noise leads to over-parameterized CNNs surpassing ResNets by 5\% in test accuracy and even more in test losses.

\section{Summary}

In summary, we consistently observe that test error decreases further in the over-parameterized regime compared to the under-parameterized regime, resulting in higher test accuracy and generalization performance for larger models. This observation contradicts the traditional bias-variance trade-off in statistical intuition. With the introduction of external label noise, the generalization performance will be correspondingly damaged, resulting in an increase in test error and forming a stronger error peak at an interpolation threshold that shifts rightwards. This observation aligns with previous research \citep{nakkiran2021deep} and coincides with an increase in test error near the interpolation threshold. 

Nevertheless, the occurrence of a test error peak at the interpolation threshold is not a universal characteristic of deep learning. Such a phenomenon is not observed in FCNNs trained on the MNIST dataset without noise. Therefore, we attribute the double descent phenomenon to the complexity inherent in the relationship between models and tasks. We speculate that it may not manifest in straightforward tasks and, conversely, might exhibit robustness in complex scenarios, such as those involving external label noise. In conclusion, the observations on the double descent (and benign over-parameterization) phenomenon align with contemporary machine learning practices and guidelines established by researchers and engineers. This intriguing phenomenon further resonates with recent research focusing on the scaling law of large language models, which has garnered significant attention in the field \cite{kaplan2020scaling}. 

While multiple studies have been raised to investigate the underlying mechanism of this phenomenon through both theoretical and empirical lenses, this study seeks to introduce several hypotheses and present evidence from perspectives including activation patterns, complexity estimation, and interpolation strategy within the hidden feature space. The aim is to offer fresh insights into this phenomenon. In the forthcoming chapter, we will begin with a hypothesis that over-parameterized networks facilitate a distinct integration of information from various representation domains, akin to the multi-head attention mechanism. This integration is believed to enhance information extraction and processing capabilities, a hypothesis we will explore in detail.

\chapter{Class-wise Activation Patterns and Correlation\label{chapter3:activation}}
The `Attention' function has attracted great `attention' and demonstrated its effectiveness across various architectures and machine learning applications. In the widely cited work on the proposal of the Transformer architecture, \citeauthor{vaswani2017attention} have proposed to use Multi-head Attention layers to replace convolutional blocks or LSTM units utilized in previous architectures for efficient information extraction and processing \citep{vaswani2017attention}. Compared to the single-head attention function, multi-head attention allows joint information from different representation spaces. Hence, as this study focuses on over-parameterized models from the viewpoint of width, we provide a hypothesis that the additional dimensions operate as distinct representation spaces, introducing novel pathways for information processing that bolster classification performance. As hidden representations undergo more independent and less correlated information processing, they can learn more robust features, facilitating improved categorization and enhancing overall generalization performance.

\section{Methodology}

We explore our hypothesis concerning the differentiation among classes, positing that larger networks handle class-specific information autonomously, resulting in reduced overlap and a more effective classification process. Specifically, larger models are expected to display more distinct activation patterns for each class compared to smaller models. In this study, the activation pattern for different classes is illustrated through the \textbf{Class Activation Matrices (CAMs)}. CAM is obtained by computing the mean of \textbf{Activation Matrices} with output ($argmax$ under classification) corresponding to each class. 

Before detailing the precise computational equation for the Activation Matrices and the CAMs, we introduce certain notations and definitions concerning the class-specific activations. We consider L-layer neural networks, with activations on each layer represented by $f_l(X)$, for $l = [0, . . . , L]$. The experiments are conducted on the training set, $(x_1, y_1),\ldots,(x_N, y_N))$ of size N. The division of the training set into class-wise categories is determined by the prediction of the classifier layer, specifically identified as the \textbf{argmax} of $f_L(x_i)$. Thus, training data categorized for each class are denoted as $\mathbb{C}_j$, for $j = [0-9]$ and all $x_i$ have $\textbf{argmax}(f_{L}(x_i)) = j$. This categorization relies on the premise that we are analyzing the internal activations of the neural network. Therefore, we utilize the network's predictions to understand the underlying factors influencing its decision-making process. 

We now start to introduce the definition and computation procedure of the CAM. We first define the Activation Matrix between layer $f_{l}$ of $n$ neurons and $f_{l+1}$ of $m$ neurons by $f_{l}(X) \cdot f_{l+1}(X).T$ of shape $n \times m$, as a product of two vectors with transpose operation on the latter to retain information about each connection between the two layers for $l in [0-L-1]$. Next, the CAM of each class $\mathbb{C}_j$ for $j \in [0-9]$ between layer $f_{l}$ and $f_{l+1}$ is given by:
\begin{equation}
    \mathrm{CAM}(f_{l+1}(\mathbb{C}_j)) = \mathbb{E} \left[ f_{l}(x_i) \cdot f_{l+1}(x_i).\mathrm{T}, \quad f_L(x_i) = j \right]
\end{equation}
By defining CAM, experiments are conducted to validate the hypothesis that the activation patterns for different classes exhibit a greater degree of isolation and decreased similarity in over-parameterized networks compared to under-parameterized ones. \textbf{Cosine Similarity} is utilized to determine the similarity between each CAM and with the mean of all pairs of distinct classes reported. The experiment results are presented in the subsequent section.

\section{Experiment Results}

In this section, initially, heatmaps depicting activation similarities between classes of an FCNN example are displayed across various model scales to facilitate an understanding of the proposed CAMs and feature extraction mechanisms in neural networks. Next, line graphs depicting the change in mean similarities among all CAMs with increasing model scales are presented to examine the correctness of our hypothesis. 

\subsection{Heatmap of CAM Similarities}

\begin{figure*}[htbp]
\begin{center}
\begin{tabular}{c}
    \includegraphics[width=1.1\textwidth]{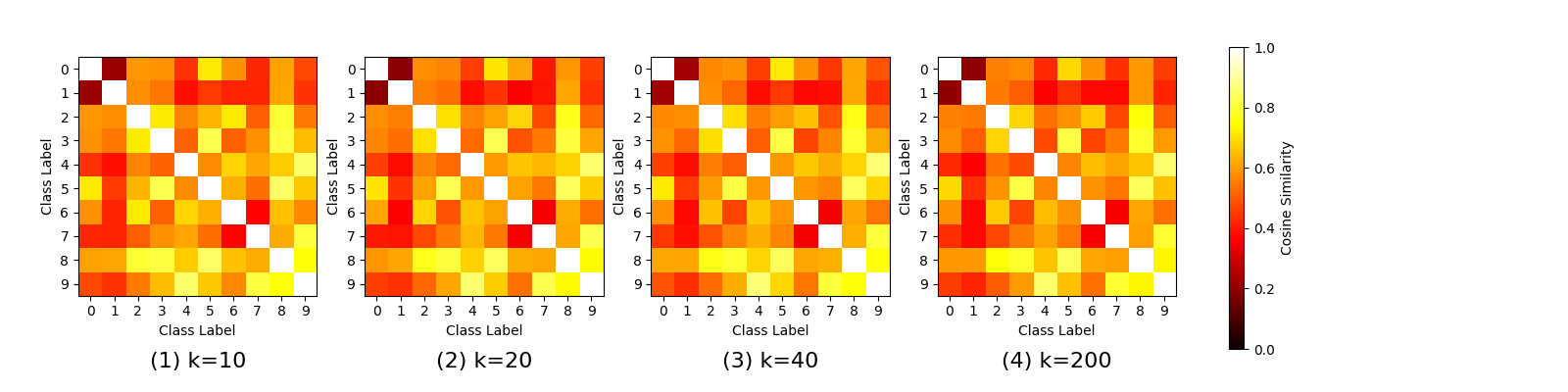} \\
    \includegraphics[width=1.1\textwidth]{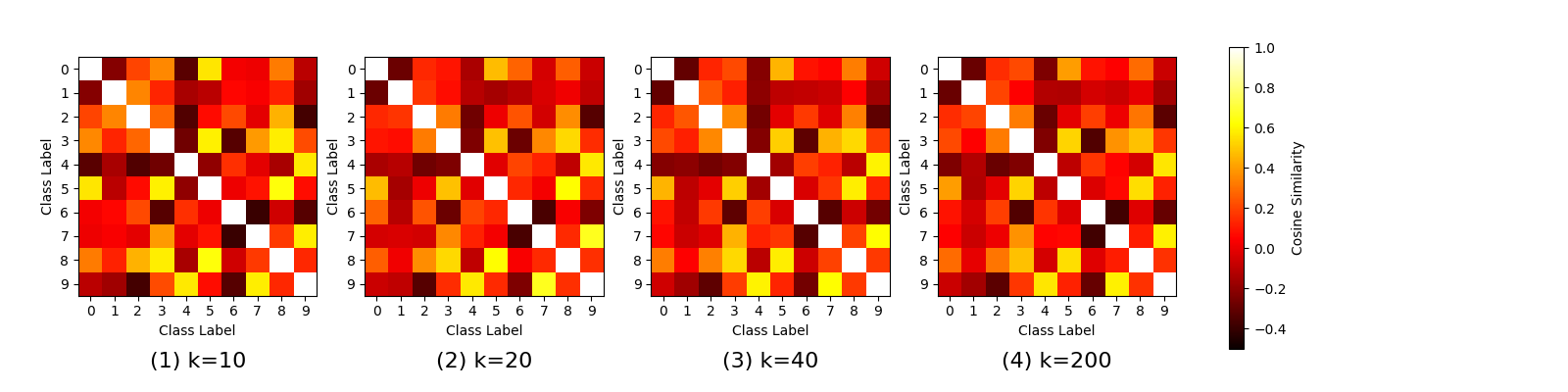} \\
\end{tabular}
\end{center}
\caption{\label{fig:CAM-Heatmap-MNIST}
    The Heatmap of the cosine similarities between every pair of the computed CAMs, drawn from one two-layer FCNN trained on MNIST with no label noise introduced as an example. Four heatmaps are selected representing stages of model parameterization (Hidden Units $k \in [10, 20, 40, 200]$). The first row of heatmaps demonstrated the similarities between the Input Layer (layer $f_0$) and the Hidden Layer (layer $f_1$). The second row of heatmaps demonstrated the similarities between the Hidden Layer (layer $f_1$) and the Output Layer (layer $f_2$). The half bottom-left section of the heatmap matrix was left as 0 and left blank. 
}
\end{figure*}

An example of the similarities between each CAM on an FCNN trained on MNIST with no label noise is presented in Figure \ref{fig:CAM-Heatmap-MNIST} across four model scales, namely $k \in [10, 20, 40, 200]$. The four different scales denote four distinct stages of parameterization: under-parameterization, interpolation threshold, over-parameterization and deep over-parameterization. While the heatmaps may not provide a clear overview of the trend or facilitate comparison between different models, we observe a mostly consistent colour pattern in the first row. However, in the second row, the colour becomes deeper with increasing neuron size, representing an increased independence between classes. 

We can also observe prominent deep colours between several pairs of classes, such as 6 and 7, 4 and 2, 3 and 6, and 2 and 9, which are all visually distinct in their components when written. Conversely, the pair of classes exhibiting less prominent colours, such as 3 and 8, 5 and 8, 4 and 9, and 7 and 9, share some components and demonstrate greater similarity compared to other pairs of numbers. These observations suggest interpretable conclusions: neural networks, when learning to classify, can discern the abstract structure of each number (class) and encourage sparsity between distinct patterns. In the future, leveraging the potential significance of activation patterns is anticipated to enhance representation learning and promote interpretable studies in deep learning.

\subsection{Mean Similarities of CAMs}

\begin{figure}[ht]
\begin{center}
\begin{tabular}{ccc}
    \begin{minipage}[b]{0.3\textwidth}
        \includegraphics[width=\textwidth]{images/Experiments/MNIST-FCNN-Epochs=4000-p=0-U.png}
    \end{minipage} &
    \begin{minipage}[b]{0.3\textwidth}
        \includegraphics[width=\textwidth]{images/Experiments/MNIST-FCNN-Epochs=4000-p=10-U.png}
    \end{minipage} &
    \begin{minipage}[b]{0.3\textwidth}
        \includegraphics[width=\textwidth]{images/Experiments/MNIST-FCNN-Epochs=4000-p=20-U.png}
    \end{minipage} \\
    
    \begin{minipage}[b]{0.3\textwidth}
        \includegraphics[width=\textwidth]{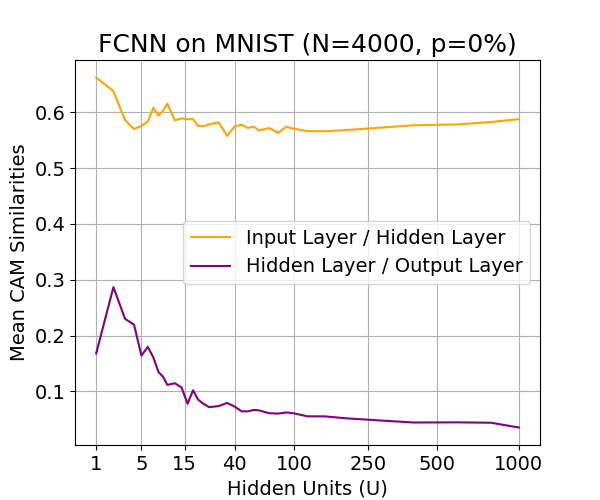}
    \end{minipage} &
    \begin{minipage}[b]{0.3\textwidth}
        \includegraphics[width=\textwidth]{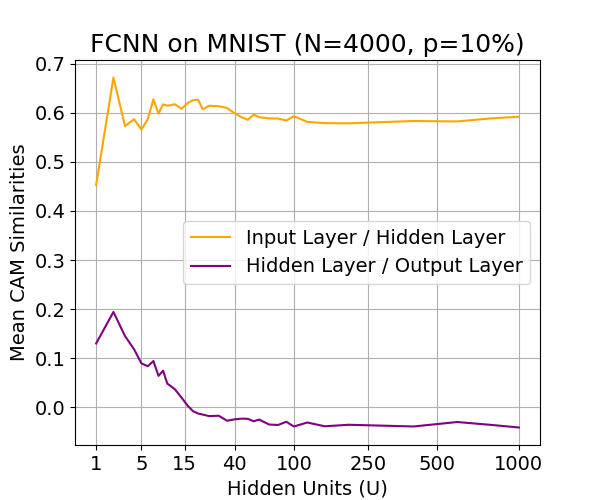}
    \end{minipage} &
    \begin{minipage}[b]{0.3\textwidth}
        \includegraphics[width=\textwidth]{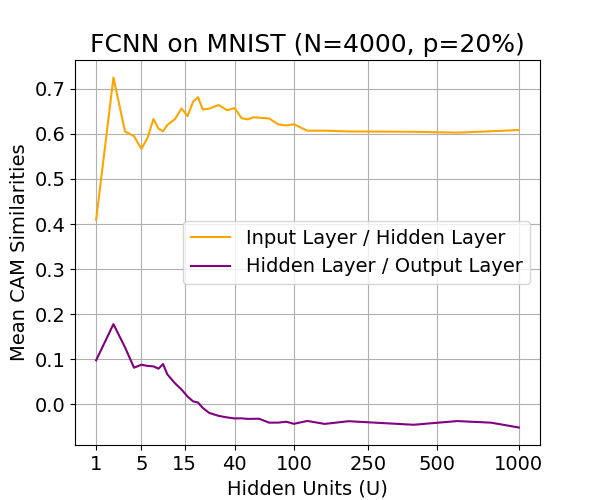}
    \end{minipage} \\
    \fontsize{8}{8}\selectfont (a) FCNN on MNIST ($p = 0\%$) &
    \fontsize{8}{8}\selectfont (b) FCNN on MNIST ($p = 10\%$) &
    \fontsize{8}{8}\selectfont (c) FCNN on MNIST ($p = 20\%$) \\
\end{tabular}
\end{center}
\caption{\label{fig:CAM-FCNN-MNIST}
    The phenomenon of double descent on two-layer FCNNs trained on MNIST $(N=4000)$, under varying explicit label noise ratios of $p = [0\%, 10\%, 20\%]$ and the mean similarities among all Class Activation Matrices (CAMs). The experimental results are placed at the top for easier comparison. The yellow line denotes the activation similarities between the input layer (layer $f_0$) and the hidden layer (layer $f_1$), whereas the purple line signifies the similarities between the hidden layer (layer $f_1$) and the output layer (layer $f_2$).
}
\end{figure}

Figure \ref{fig:CAM-FCNN-MNIST} presented the mean similarities over CAMs of all classes over the number of hidden units on FCNN trained on MNIST, with the experiment results provided for better interpretation. We will first inspect the Input-Hidden injections w.r.t. increase in the model width. In Figure \ref{fig:CAM-FCNN-MNIST}(a), we observe that no notable changes are introduced when there is no label noise with mean cosine similarities of 0.6. The yellow line exhibits a general pattern of decreasing followed by increasing values across the model range, with fluctuations that do not show a distinct trend in the first half. In Figure \ref{fig:CAM-FCNN-MNIST}(b), when 10\% label noise is introduced, we observe a notable off-range value with model width $k=2$, while similarities generally increase beyond the interpolation threshold. Subsequently, they follow a similar pattern of decrease followed by an increase. Lastly, in Figure \ref{fig:CAM-FCNN-MNIST}(c) with 20\% label noise, a more pronounced peak is noticeable at $k=2$, accompanied by a stronger peak at the interpolation threshold. The numerical values of similarities then decrease to 0.6 in the over-parameterized regime. Neglecting the extremely small models with width $k<5$, we can see the trend in mean similarities aligns with the double descent phenomenon to some extent. With the introduction of additional label noise, the class-wise input representation exhibited higher similarities, particularly noticeable at the interpolation threshold, before dropping to a similar value in the over-parameterized regime. This observation is also consistent with our hypothesis that the model should exhibit reduced similarities across CAMs.

Next, we will examine the Hidden-Output layer injection, also referred to as the classifier. We may observe that if again ignoring models with width $k=1,2$, the mean similarities between all CAMs exhibit a constantly decreasing trend in Figure \ref{fig:CAM-FCNN-MNIST}(a) with numerical values of 0.3 to 0, suggesting a decoupling of class-specific activation patterns during the expansion of model width. While the Input-Hidden layer is commonly termed the representation layer, we can infer that implicit activation sparsity between class patterns emerges solely in the classifier, while the representation layer encapsulates data on a universal basis. Nevertheless, with additional label noise introduced in Figure \ref{fig:CAM-FCNN-MNIST}(b, c) the overall similarities dropped overall by 0.1 and 0.12, and remained consistent after the interpolation threshold. This could be attributed to models suffering from increased noise information and randomness in training data while lacking sufficient reasons to believe. In conclusion, the experiment results indicate that the mean similarities between all CAMs tend to increase at the representation layer and decrease at the classification layer before the interpolation threshold, and remain mostly consistent afterwards. The tendency will be enhanced by additional label noise introduced on the training dataset. 

\begin{figure}[ht]
\begin{center}
\begin{tabular}{ccc}
    \begin{minipage}[b]{0.3\textwidth}
        \includegraphics[width=\textwidth]{images/Experiments/CIFAR-10-CNN-Epochs=200-p=0-U.png}
    \end{minipage} &
    \begin{minipage}[b]{0.3\textwidth}
        \includegraphics[width=\textwidth]{images/Experiments/CIFAR-10-CNN-Epochs=200-p=10-U.png}
    \end{minipage} &
    \begin{minipage}[b]{0.3\textwidth}
        \includegraphics[width=\textwidth]{images/Experiments/CIFAR-10-CNN-Epochs=200-p=20-U.png}
    \end{minipage} \\
    
    \begin{minipage}[b]{0.3\textwidth}
        \includegraphics[width=\textwidth]{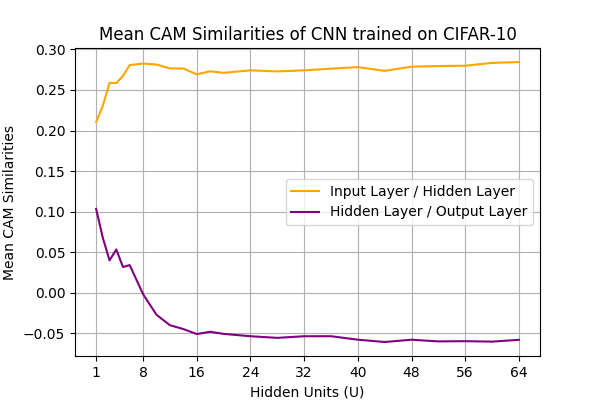}
    \end{minipage} &
    \begin{minipage}[b]{0.3\textwidth}
        \includegraphics[width=\textwidth]{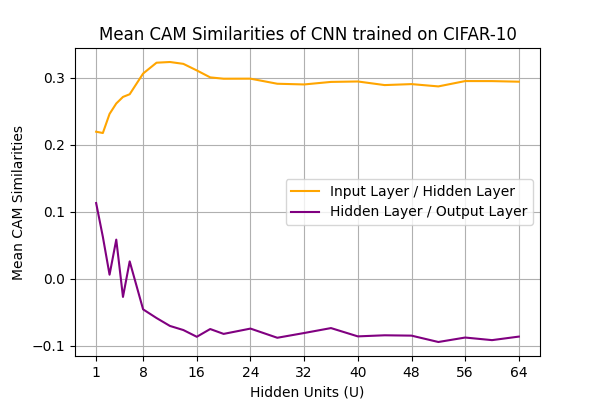}
    \end{minipage} &
    \begin{minipage}[b]{0.3\textwidth}
        \includegraphics[width=\textwidth]{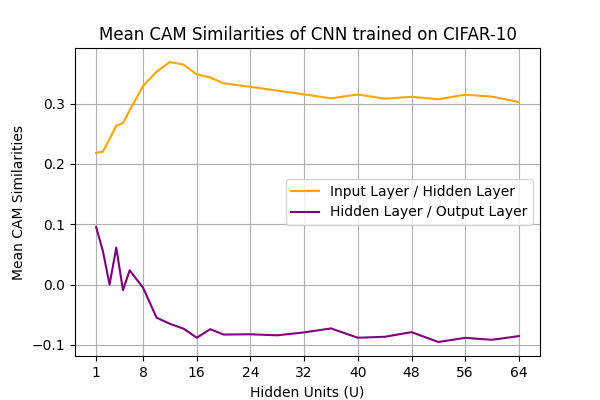}
    \end{minipage} \\
    \fontsize{8}{8}\selectfont (a) CNN on CIFAR-10 ($p = 0\%$) &
    \fontsize{8}{8}\selectfont (b) CNN on CIFAR-10 ($p = 10\%$) &
    \fontsize{8}{8}\selectfont (c) CNN on CIFAR-10 ($p = 20\%$) \\
\end{tabular}
\end{center}
\caption{\label{fig:CAM-CNN-CIFAR-10}
    The phenomenon of double descent on five-layer CNNs trained on CIFAR-10 ($N=50000$), under varying label noise ratios of $p = [0\%, 10\%, 20\%]$ and the mean similarities among all Class Activation Matrices (CAMs). The experimental results are placed at the top for easier comparison. The yellow line denotes the activation similarities between the input layer (layer $f_0$) and the hidden layer (layer $f_4$), whereas the purple line signifies the similarities between the hidden layer (layer $f_4$) and the output layer (layer $f_5$).
}
\end{figure}

\begin{figure}[ht]
\begin{center}
\begin{tabular}{ccc}
    \begin{minipage}[b]{0.3\textwidth}
        \includegraphics[width=\textwidth]{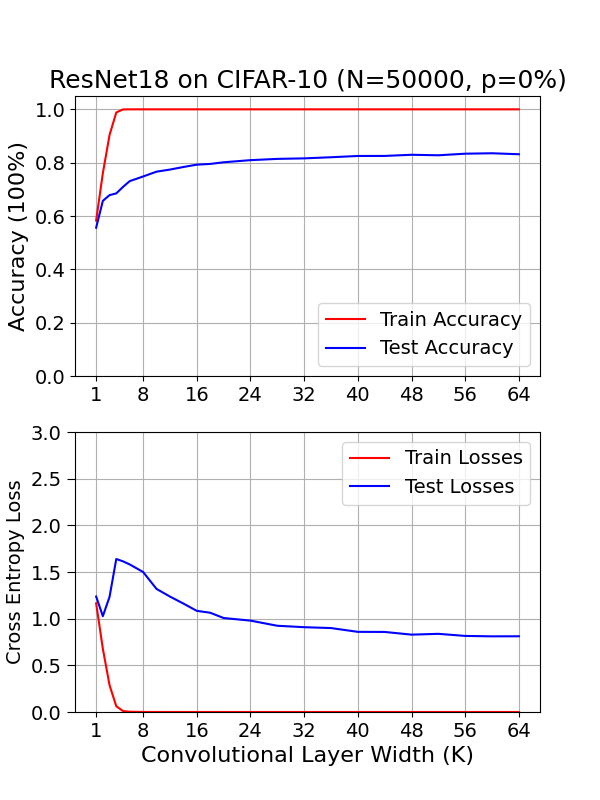}
    \end{minipage} &
    \begin{minipage}[b]{0.3\textwidth}
        \includegraphics[width=\textwidth]{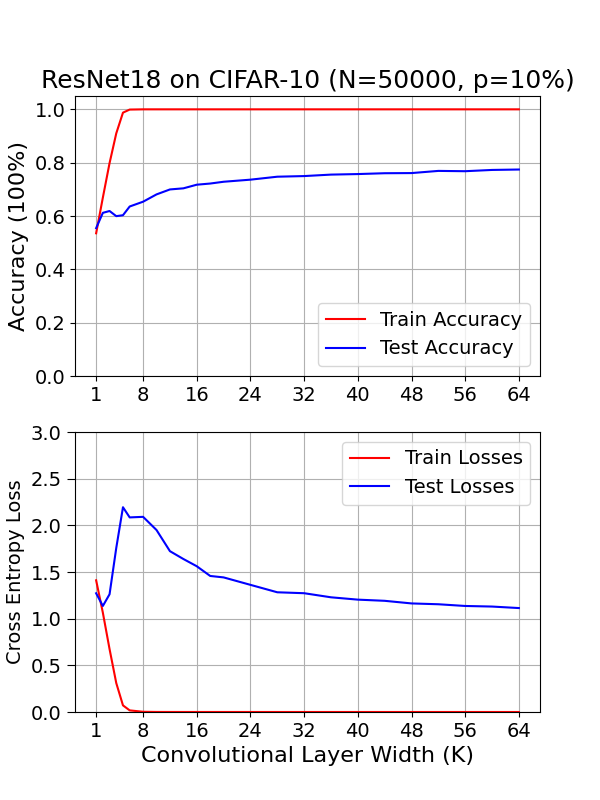}
    \end{minipage} &
    \begin{minipage}[b]{0.3\textwidth}
        \includegraphics[width=\textwidth]{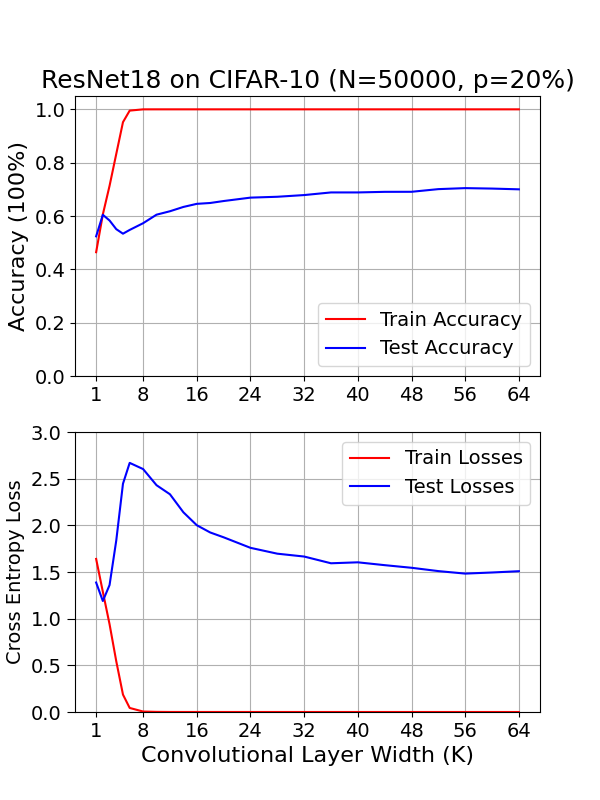}
    \end{minipage} \\
    
    \begin{minipage}[b]{0.3\textwidth}
        \includegraphics[width=\textwidth]{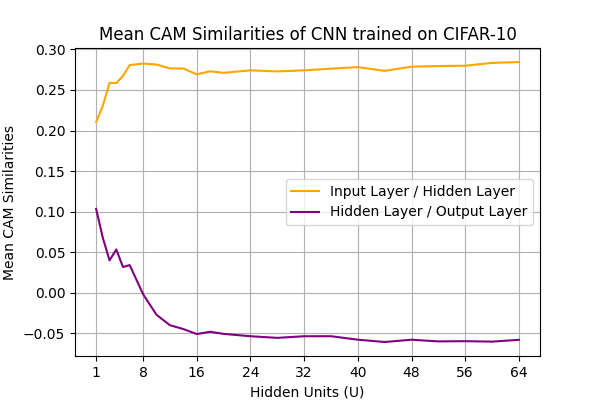}
    \end{minipage} &
    \begin{minipage}[b]{0.3\textwidth}
        \includegraphics[width=\textwidth]{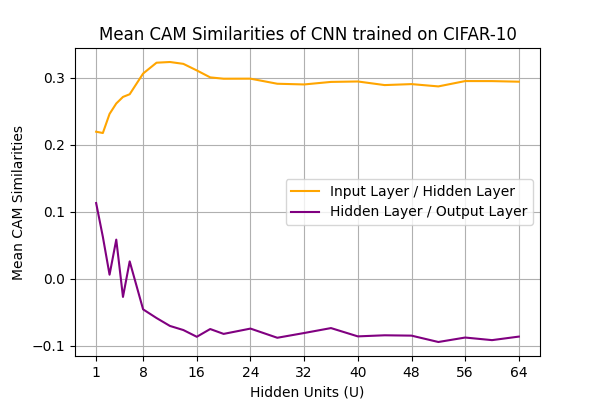}
    \end{minipage} &
    \begin{minipage}[b]{0.3\textwidth}
        \includegraphics[width=\textwidth]{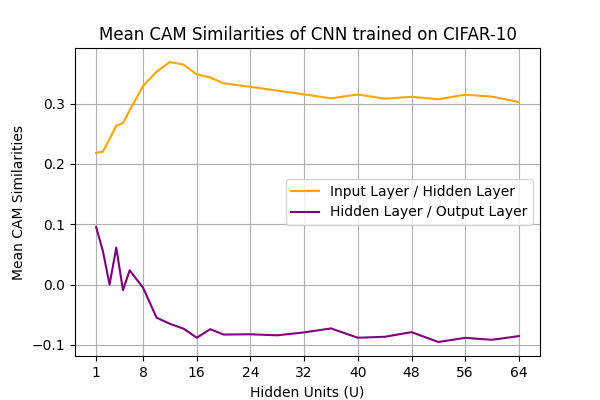}
    \end{minipage} \\
    \fontsize{8}{8}\selectfont (a) ResNet on CIFAR-10 ($p = 0\%$) &
    \fontsize{8}{8}\selectfont (b) ResNet on CIFAR-10 ($p = 10\%$) &
    \fontsize{8}{8}\selectfont (c) ResNet on CIFAR-10 ($p = 20\%$) \\
\end{tabular}
\end{center}
\caption{\label{fig:CAM-ResNet-CIFAR-10}
    The phenomenon of double descent on ResNet18s trained on CIFAR-10 $(N=50000)$, under varying label noise of $p = [0\%, 10\%, 20\%]$ and the mean similarities among all Class Activation Matrices (CAMs). The experimental results are placed at the top for easier comparison. The yellow line denotes the activation similarities between the input layer (layer $f_0$) and the hidden layer (layer $f_4$), whereas the purple line signifies the similarities between the hidden layer (layer $f_4$) and the output layer (layer $f_5$).
}
\end{figure}

Secondly, we will inspect the same experiment conducted on CNNs and ResNet18s trained on CIFAR-10 in Figure \ref{fig:CAM-CNN-CIFAR-10} and \ref{fig:CAM-ResNet-CIFAR-10}, where we will see our conclusions drawn from FCNNs is mostly applicable. For the Input-Hidden layer injection, among the three test cases with different label noise ratios, we can see a clear similarity peak between the bottom of the `U-shaped' test loss curve and the interpolation threshold, and the increased noise ratio enhanced the peak and overall similarity. Due to the four convolutional layer's superior ability to capture representations for each class, the similarities ranged from 0.2 to 0.4, significantly lower than FCNN's representation layer of 0.6. For the Hidden-Output classifier layer, the similarities value between CNNs and ResNet18s are comparable to those of FCNNs: the similarities decrease before reaching the interpolation threshold and remain consistent thereafter; increased label noise led to a relative drop in similarity. Fluctuations are observed in the under-parameterized regime, although the overall trend remains evident. Based on experiments conducted with CNNs and ResNet18s, we can infer that the class-wise similarities in the classifier layer may somehow mirror the training outcomes, whereas the class-wise similarities in the representation layer may align with the testing outcomes.

\section{Summary}

In summary, by drawing from the idea of multi-head attention and exploring the activation patterns that appeared in the neural networks, several insights are revealed: 1. The class patterns are more pronounced in the classification layer than the representation layer; 2. In the representation layer, a similarity peak is formed before the interpolation threshold and decreases mostly afterwards; 3. Neural model tends to establish more independent representations for each class for classification under intrinsic model dimension; 4. CAMs established in the classifier facilitate some interpretable patterns, proving that neural networks can learn abstract features corresponding to each image class. These conclusions validate our hypothesis outlined at the beginning of the chapter: wider neural networks demonstrate distinct pathways for features associated with each class, potentially enhancing robust feature extraction through reduced correlation between class-wise activations. 

However, it's important to acknowledge that although evidence indicates that over-parameterized neural networks demonstrate more discernible activation patterns for individual classes than under-parameterized ones, we still need to explore why increased model dimensionality leads to improved results, even in the over-parameterized regime, where the similarities in CAMs remain largely consistent. Additionally, it is puzzling why neural models with increased dimensionality do not suffer from overfitting, particularly with growing label noise. It remains a challenge to the conventional statistical belief that increasing model dimensionality should introduce greater complexity and additional variance. Contrary to this expectation, we observe that over-parameterization further reduces test error in practice. Thus, in the upcoming chapter, we will examine the model complexity of these class-wise activation patterns. An estimation methodology is proposed by drawing intuition from the Rademacher Complexity and sub-sampling to elucidate if increasing dimension brings higher complexity in implicit activations.

\chapter{Estimating the Class-wise Activation Complexity\label{chapter4:complexity}}
In deep learning theory, the term `model complexity' may refer to two different meanings and are captured by the notions of model expressive capacity and model effective complexity, respectively \citep{hu2021model}. Expressive capacity measures the ability of deep learning models to approximate complex problems, while effective model complexity reflects the complexity of the functions represented by deep models with specific parameterization \citep{raghu2017expressive}. In traditional statistical intuition, model complexity increases as the number of parameters increases in the model, which should imply higher variance and poor generalization performance. Yet, the presence of the benign over-parameterization phenomenon implies either robustness in complex neural networks against overfitting or a disconnect between the number of parameters and complexity. When presenting the empirical studies on deep double descent, \citet{nakkiran2021deep} also introduced a measure of the effective model complexity (EMC) as the maximum number of samples the model can attain zero training error. It has been argued that the general double descent phenomenon can be categorized as a function of EMC \citep{nakkiran2021deep}.

Differing from the proposed EMC notion, this chapter delves into a different complexity measure of neural networks on class-wise activations. The proposed data-dependent methodology is extended from the richness of the activations, drawing from the concept of Rademacher complexity. In computational learning theory, Rademacher complexity, named after Hans Rademacher, measures the richness of a class of real-valued functions to a probability distribution \cite{bartlett2002rademacher}. Although Rademacher complexity has been extensively explored and applied in theoretical learning theory studies to gauge model complexity, its practical application in real-world empirical studies is challenged by the exponential computational complexity it entails over the number of testing samples. 

Therefore, this study introduces an estimation methodology incorporating sub-sampling and randomness to offer a rough assessment of the diversity of a class's output. The aim of introducing this methodology is to provide a means for comparing the richness of all class-wise activation in deep learning models, complementing the discussion from the previous chapter. The estimated richness served as an indirect measure of model complexity from an internal perspective. We hypothesized that for a classifier, the complexity of each class-wise activation pattern is more representable as the effective complexity than data dependent on the full data distribution. We posited that for a classifier, the effective complexity is better characterized by the complexity of each class-wise activation pattern than derived from the entire input data distribution. By combining discussions from the last chapter, we further hypothesized that neural models exhibit more distinguishable and simpler activation patterns for each class w.r.t. increasing size (model width). This hypothesis likely explains the decrease in variance and the enhanced generalization performance against overfitting.

\section{Methodology}

We will reintroduce some notations before presenting the estimation methodology of model complexity. We considered L-layer neural networks, with activations on each layer represented by $f_l(X)$, for $l = [1, . . . , L]$. Unseen testing data are utilized as inputs, with the image set categorized for each class denoted as $\mathbb{C}_j$, for $j \in [0-9]$ and all $x_i$ have $\textbf{argmax}(f_{L}(x_i)) = j$. 

We now start to estimate the richness of the hidden representation of all samples predicted as each class. Testing images $\mathbb{C}_j$ predicted as the same class is partitioned into subsets of 20 images per group. For each group, 50 Rademacher variables of size 20 are randomly sampled and their dot product is calculated with the 20 image activations at the hidden layer. The maximum value is computed over the mean of each of the 50 samples, and the mean over the maximum values of all groups is adopted as the estimate of the richness of class activations, further referred to as their complexity. Below is a mathematical equation provided as a reference to the estimate:
\begin{equation}
    {\mathrm{Richness}(f_{L-1}(\mathbb{C}_j))} = \mathbb{E} \left[ \sup \left( \mathbb{E} [ \sigma_i \cdot f_{L-1}(x_i) ], \hspace{4pt} i \in [20k+1, 21k], \hspace{4pt} \sigma_i \in [-1, 1] \right) \right] \nonumber,
\end{equation}
where $k$ referred to the group number, $\sigma_i$ referred to the Rademacher variable, $j$ corresponds to the class number. The richness of the neural network's implicit activation is then given by:
\begin{equation}
    \mathrm{Richness}(f_{L-1}) = \mathbb{E} \left[ \mathrm{Richness}(f_{L-1}(C_j)), \hspace{4pt} j \in [0-9] \right]
\end{equation}

The proposed methodology presented here addresses the challenge of sampling all Rademacher variables for every data point by employing subsampling on the dataset. Subsequently, Rademacher variables are sampled over these subsamples of data, and the mean over the maximum of all groups is computed to provide a rough estimate of the relative richness among hidden activations of each class. This approximate estimation is anticipated to enable improved comparisons between deep learning models and enhance understanding of activations' richness. Concluded by \citet{hu2021model}, there are four key aspects: model framework, model size, optimization process and data complexity that affect model complexity. It's worth noting that in our experiment setup, the framework, training dataset, and optimization process are consistent for each test case, with only the model size increasing. Therefore, the experiment results should solely pertain to the model size and subsequently to benign over-parameterization. These experiment results are presented in the next section of all trained neural networks.

\section{Experiment Results}

\begin{figure}[ht]
\begin{center}
\begin{tabular}{ccc}
    \begin{minipage}[b]{0.3\textwidth}
        \includegraphics[width=\textwidth]{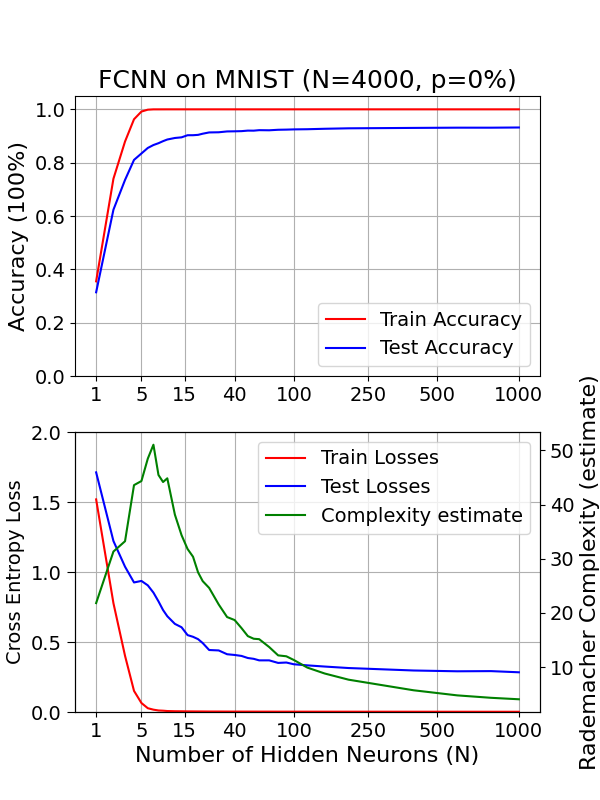}
    \end{minipage} &
    \begin{minipage}[b]{0.3\textwidth}
        \includegraphics[width=\textwidth]{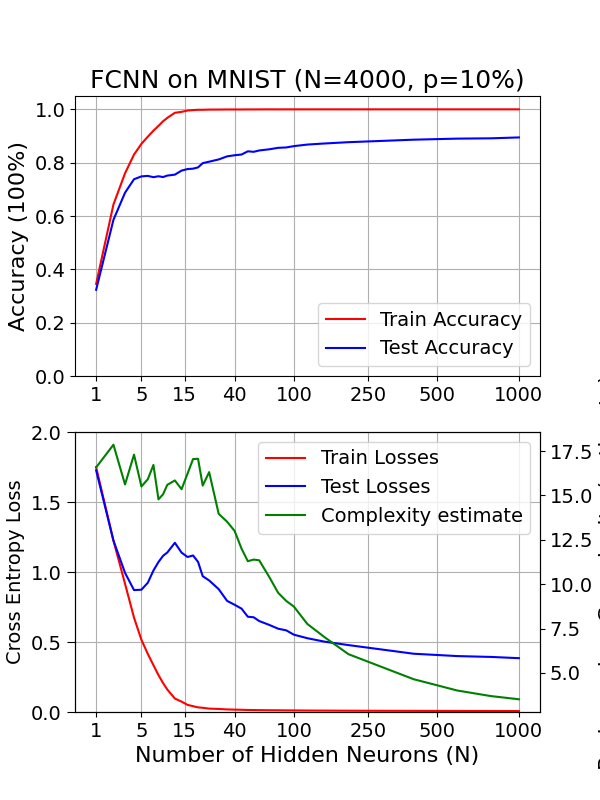}
    \end{minipage} &
    \begin{minipage}[b]{0.3\textwidth}
        \includegraphics[width=\textwidth]{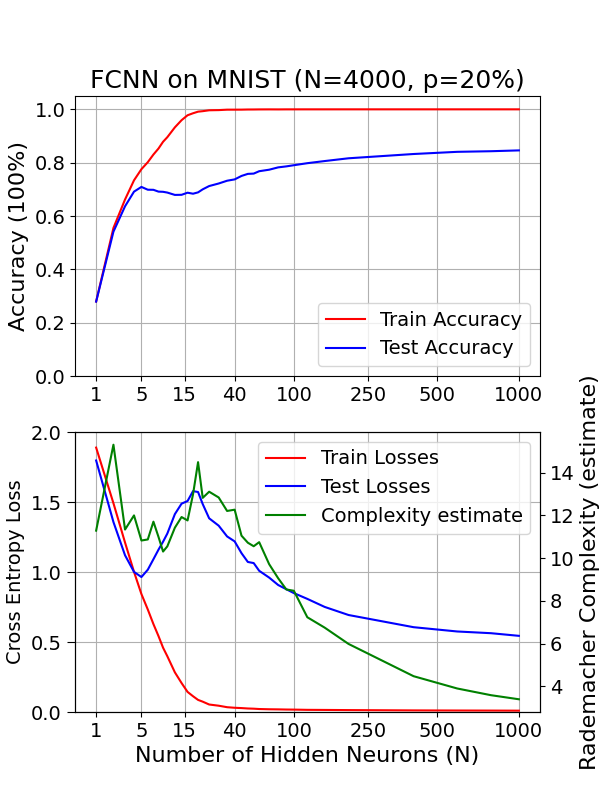}
    \end{minipage} \\
`   \fontsize{8}{8}\selectfont (a) FCNN on MNIST $(p=0\%)$ & 
    \fontsize{8}{8}\selectfont (b) FCNN on MNIST $(p=10\%)$ & 
    \fontsize{8}{8}\selectfont (c) FCNN on MNIST $(p=20\%)$ \\
\end{tabular}
\end{center}
\caption{\label{fig:Rade-MNIST-FCNN}
    The phenomenon of double descent on two-layer FCNNs trained on MNIST $(N=4000)$, under varying explicit label noise ratios of $p = [0\%, 10\%, 20\%]$ and the mean estimated richness of class-wise activations. The peak in estimated richness is observed also at the peak in test loss and at the interpolation threshold. 
}
\end{figure}

From Figure \ref{fig:Rade-MNIST-FCNN}, we can see the estimated model richness of FCNNs trained on MNIST by class-wise activations on the training dataset. We can observe a consistent peak in richness, which occurred slightly after the peak in test loss at the interpolation threshold, across various label noise ratios. After passing the interpolation threshold, this peak in estimated complexity significantly decreases as the benign over-parameterization phenomenon occurs w.r.t. increase in model width. This observation is thought to be closely related to the latter phenomenon and may offer insights into the idea that increasing model width and parameterization in the end decreases the complexity in implicit activation patterns. 

Conversely, the trend in richness before reaching the interpolation threshold appears to be rather unpredictable across different noise ratios, exhibiting an increasing pattern with no label noise, a fluctuated high complexity line under 10\% label noise and a `U-shaped' curve aligning with the test loss under 20\% label noise. One explanation is that the increased label noise creates difficulties for under-parameterized models fitting the data and thus increases complexity in activation, though we may observe that the estimated value of all models at different scales decreases w.r.t the increase in noise ratio. While the absolute value of this estimated richness holds no significance in this context, the decrease in relative richness with increasing noise is an intriguing observation that merits further investigation. One hypothesis suggests that the increased label noise functions as adversarial training samples, contributing to the promotion of robustness in neural network activations and consequently reducing the complexities. Further discussion on the role of the added label noise will be addressed in the next chapter.

\begin{figure}[ht]
\begin{center}
\begin{tabular}{ccc}
    \begin{minipage}[b]{0.3\textwidth}
        \includegraphics[width=\textwidth]{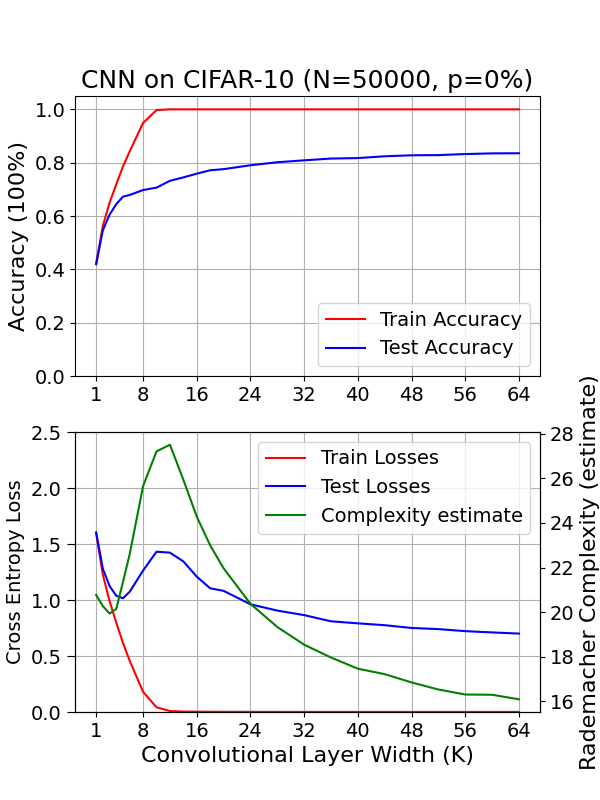}
    \end{minipage} &
    \begin{minipage}[b]{0.3\textwidth}
        \includegraphics[width=\textwidth]{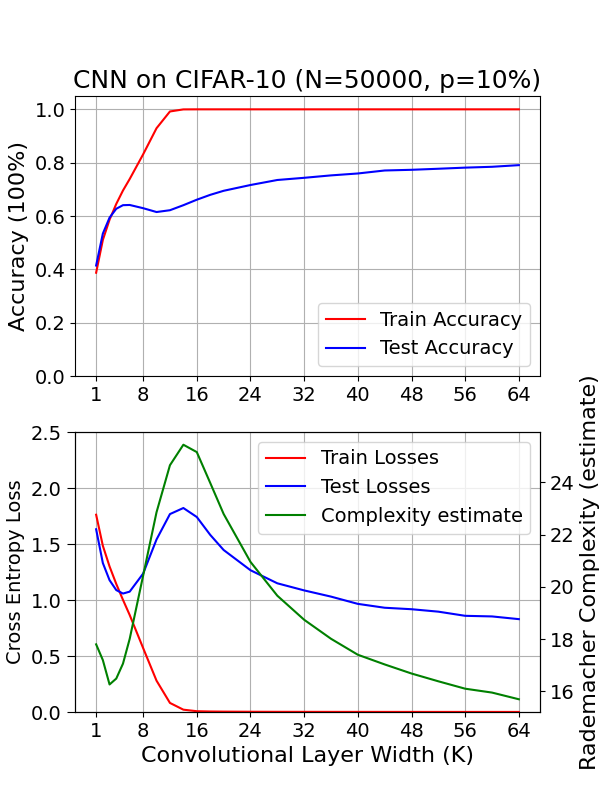}
    \end{minipage} &
    \begin{minipage}[b]{0.3\textwidth}
        \includegraphics[width=\textwidth]{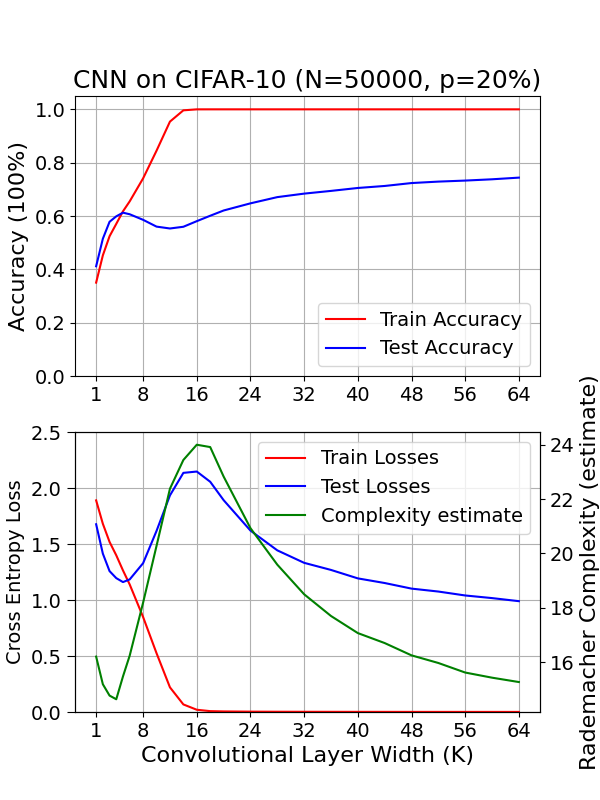}
    \end{minipage} \\
    \fontsize{8}{8}\selectfont (a) CNN on CIFAR-10 $(p=0\%)$ & 
    \fontsize{8}{8}\selectfont (b) CNN on CIFAR-10 $(p=10\%)$ & 
    \fontsize{8}{8}\selectfont (c) CNN on CIFAR-10 $(p=20\%)$ \\
\end{tabular}
\end{center}
\caption{\label{fig:Rade-CIFAR-10-CNN}
     The phenomenon of double descent on five-layer CNNs trained on CIFAR-10 ($N=50000$), under varying label noise ratios of $p = [0\%, 10\%, 20\%]$ and the mean estimated richness of class-wise activations of five-layer CNNs trained on CIFAR-10 across various widths with different label noise $p=[0\%, 10\%, 20\%]$. All of the richness curve highly aligns in tendency with the test loss curve. 
}
\end{figure}

We then move on to the experiments with five-layer CNNs and ResNet18s trained on CIFAR-10. In the experiment setups with CNNs, we can observe all richness curves under different label noises highly align in tendency with the test loss curve in Figure \ref{fig:Rade-CIFAR-10-CNN}. We may notice that the lowest point with the minimum richness precedes the bottom of the `U-shaped' partition of the test curve and the peak in maximum richness again occurs after the interpolation threshold. The decrease in richness continues consistently in the over-parameterized regime, aligning with the decrease in test error and increase in test accuracy. As the label noise ratio increases, the richness curve shifts rightwards with the interpolation threshold and overall decreases in magnitude.

\begin{figure}[ht]
\begin{center}
\begin{tabular}{ccc}
    \begin{minipage}[b]{0.3\textwidth}
        \includegraphics[width=\textwidth]{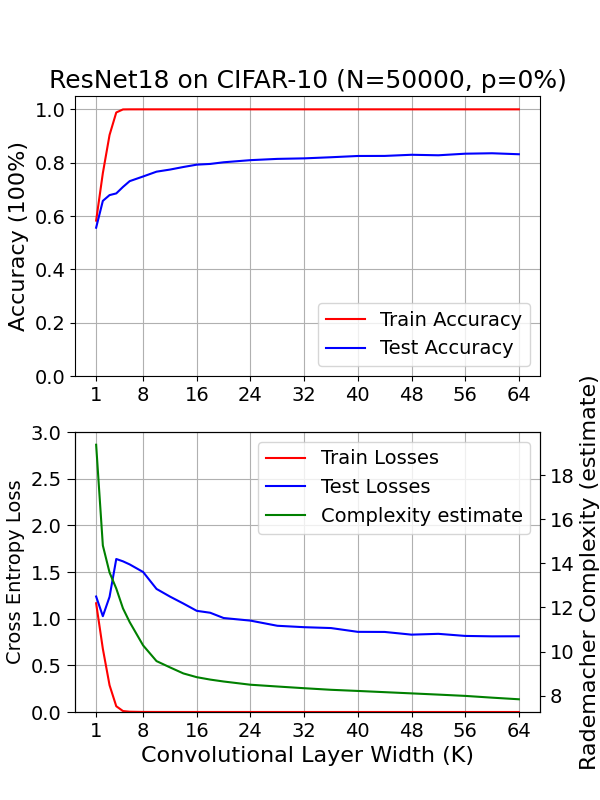}
    \end{minipage} &
    \begin{minipage}[b]{0.3\textwidth}
        \includegraphics[width=\textwidth]{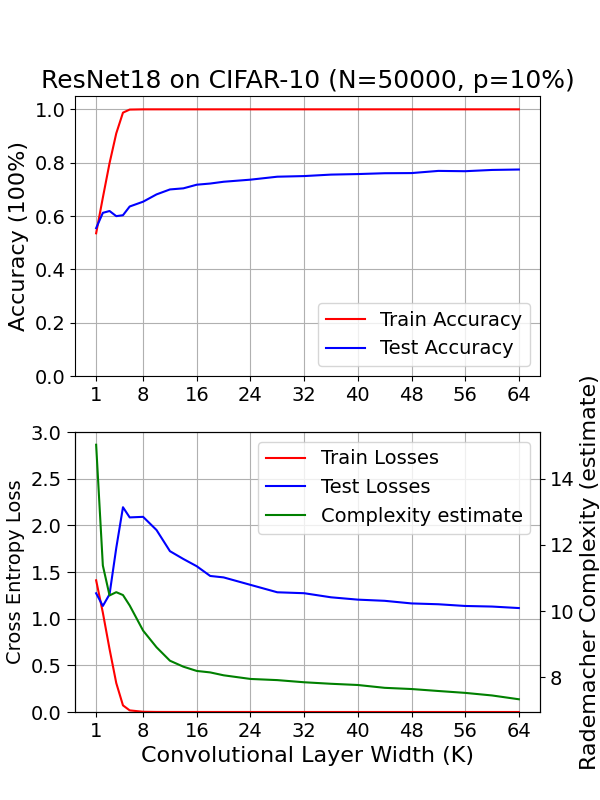}
    \end{minipage} &
    \begin{minipage}[b]{0.3\textwidth}
        \includegraphics[width=\textwidth]{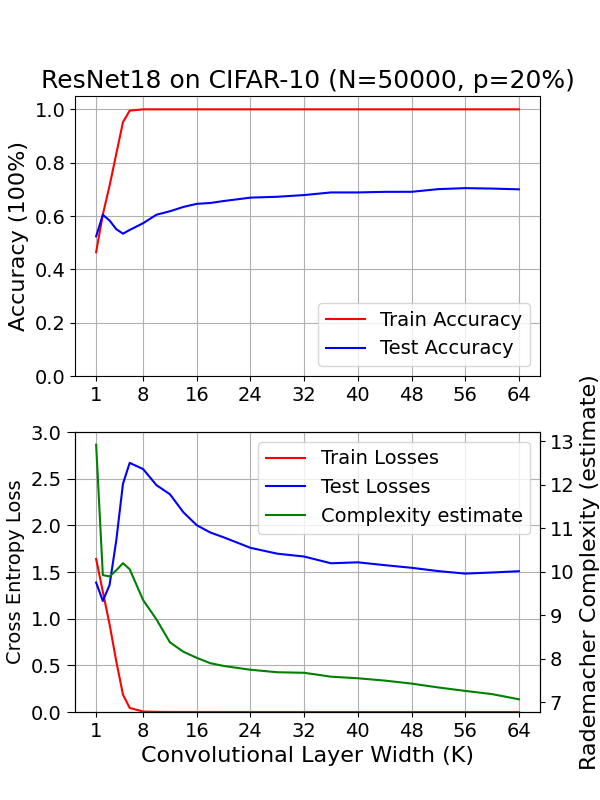}
    \end{minipage} \\
    \fontsize{8}{8}\selectfont (a) ResNet on CIFAR-10 $(p=0\%)$ & 
    \fontsize{8}{8}\selectfont (b) ResNet on CIFAR-10 $(p=10\%)$ & 
    \fontsize{8}{8}\selectfont (c) ResNet on CIFAR-10 $(p=20\%)$ \\
\end{tabular}
\end{center}
\caption{\label{fig:Rade-CIFAR-10-ResNet18}
    The phenomenon of double descent on ResNet18s trained on CIFAR-10 $(N=50000)$, under varying label noise of $p = [0\%, 10\%, 20\%]$ and the mean estimated richness of class-wise activations. All of the richness curve highly aligns in tendency with the test loss curve. 
}
\end{figure}

In the experiment setups with ResNets in Figure \ref{fig:Rade-CIFAR-10-ResNet18}, we can see when no label noise is introduced, the complexity estimate curve does not exhibit the same double descent phenomenon as the test error curve. A potential reason for this is the superior capability of ResNets significantly compressed the under-parameterized regime and erased further changes. With label noise introduced, the double descent phenomenon once again arises at the interpolation threshold, though the peak and first increase are also not significant. Nevertheless, the same conclusion of the richness/complexity decreases consistently in the over-parameterized regime still stands. The overall numerical values also decrease w.r.t. increase in label noise ratio, which remains mysterious. 

The observations in all three groups of experiment results all shed light on an explanation for benign over-parameterization, that with no regularization methodology applied and no sparsity raised in model architecture (referred to and discussed in Appendix \ref{appendix:sparsity}), the richness in activation and estimated complexity in model activations for each class functions decreases in larger models (with a potential cause of detached activation patterns discussed in the previous chapter), and contributes to reduced variance and better generalization performance.

\section{Summary}

In this chapter, we explored an empirical approach to estimating the richness and complexity of class-wise activations of neural networks, drawing inspiration. The proposed methodology differs from the EMC notion by \citet{nakkiran2021deep} and is more closely related to the concept of Rademacher complexity. Through experiments conducted on three neural architectures with MNIST and CIFAR-10 and different ratios of label noise, we observed a peak in the complexity of mean class-wise activation at the hidden layer before classification at the interpolation threshold. Passing the interpolation threshold, this richness or complexity significantly drops w.r.t increase in model width unit $k$. While the trend is less consistent before reaching the interpolation threshold, it's noteworthy that this richness still correlates to some extent with the shape of the test loss or test accuracy curve. 

In conclusion, the experimental results reveal a remarkable connection between the complexity of class-wise activations and the double descent phenomenon. It essentially confirms our hypothesis that over-parameterized networks not only exhibit more discernible class activation patterns but also share lower complexity in each pattern. This phenomenon may eventually contribute to reduced variance and enhanced generalization against overfitting. We hope our findings can offer insights into the phenomenon and encourage further exploration in the future.

Nevertheless, despite conducting experiments across various label noise ratios, which theoretically should intensify overfitting during neural network training, we have yet to investigate how this noise influences the performance gap at the interpolation threshold and how over-parameterized networks overcome this challenge. In the next and final empirical chapter, we will delve into how neural networks interpolate noisy data in the hidden feature space, and explore how various interpolation strategies can isolate noise information.

\chapter{Interpolation of Noisy Data in the Feature Space\label{chapter5:interpolation}}
In this chapter, our main focus is to investigate how the presence of stronger label noise leads to a more pronounced peak at the interpolation threshold across different neural architectures, and how over-parameterized models avoid overfitting with the presence of additional noise and still outperform under-parameterized ones. Our primary research inquiry aims to gain insights into how neural networks of various widths interpolate the noisy labelled data in the hidden feature space.

\section{Methodology}

Based on our observation of the benign over-parameterization phenomenon, we are safe to assume that test images, akin to training images mislabeled, are more likely to be correctly classified by over-parameterized models. For instance, we anticipate that an optimal classifier should yield accurate predictions for unseen images, even if it was trained on similar images with an adversarial label. Thus, we further hypothesize that the closest neighbours of mislabeled training images are correctly classified for a model with better generalization performance. To validate our hypothesis, we adopted the $k$-nearest neighbour algorithm with cosine similarity to interpret the relative locations of clean and noisy labelled data in the learned feature space. This methodology allows us to delineate the prediction strategy indirectly and subsequently compare our findings with the overall generalization performance of these pre-trained models.

We consider L-layer neural networks, while representations on each layer are represented by $f_l(X)$, for $l = [1, . . . , L]$. Our investigation centred on the penultimate layer representations before classification denoted as $f_{L-1}(X)$. Acknowledging the difficulties of interpreting high-dimensional feature spaces, our approach mainly entails characterizing the relative positioning of training data points. This involves establishing a connection between the hidden features of noisy data points in the learned feature space and their $k$-nearest neighbours within the complementary subset of clean data. So we calculate the prediction accuracy $P$ of mislabeled training data and the majority of its nearest neighbours in feature space are in the same class: 
\begin{equation}
    P = \frac {\sum^{m}_{i=1} [y_i = M(f_{L-1}(x_{i,1}),\ldots,f_{L-1}(x_{i,k}))]} {m},
\end{equation}
In this context, $(x_1, y_1),\ldots,(x_m, y_m)$ denotes the subset of noisy training data with original labels, and $(x_{i,1}, y_{i,1}),\ldots,(x_{i,\hat{m}}, y_{i,\hat{m}}))$ denotes the subset of clean training data as neighbours to $x_i$ for all $i = [1,\ldots,m]$. Given the cosine similarity $S_C(A, B):= \frac{A \cdot B}{\|A\| \|B\|}$ between two representations A and B, let $f_{L-1}(x_{i,1}),\ldots,f_{L-1}(x_{i,\hat{m}})$ be a reordering of the representations of the clean training data such that $S_C(f_{L-1}(x_i), f_{L-1}(x_{i,1})) \leq \cdots \leq S_C(f_{L-1}(x_i), f_{L-1}(x_{i,\hat{m}}))$. The closest $k$ neighbours are selected, and a Majority function $M(f_{L-1}(x_{i,1}),\ldots,f_{L-1}(x_{i,k}))$ describes the procedure for making predictions based on the majority label among the k-nearest neighbours. The predictions are subsequently compared with the original labels of noisy data, and $P$ denotes the proportion of accurate predictions among all the noisy data. 

In line with our hypothesis, the prediction accuracy $P$ is indicative of the neural architecture's interpolation approach toward these noisy labelled data points. Due to the uncertain distribution of data and their representations, the $k$-NN method functions solely as an estimation technique, and our comparison focuses on variations in $P$. The empirical study results of this hidden feature space interpretation are presented in the next section.

\section{Experiment Results}

In this section, we present our experiment results with a variety of neural architectures of increasing size trained for image classification tasks on either MNIST or CIFAR-10 datasets compared to the generalization performance represented by test loss and accuracy.

\begin{figure}[ht]
\begin{center}
\begin{tabular}{ccc}
    \begin{minipage}[b]{0.3\textwidth}
        \includegraphics[width=\textwidth]{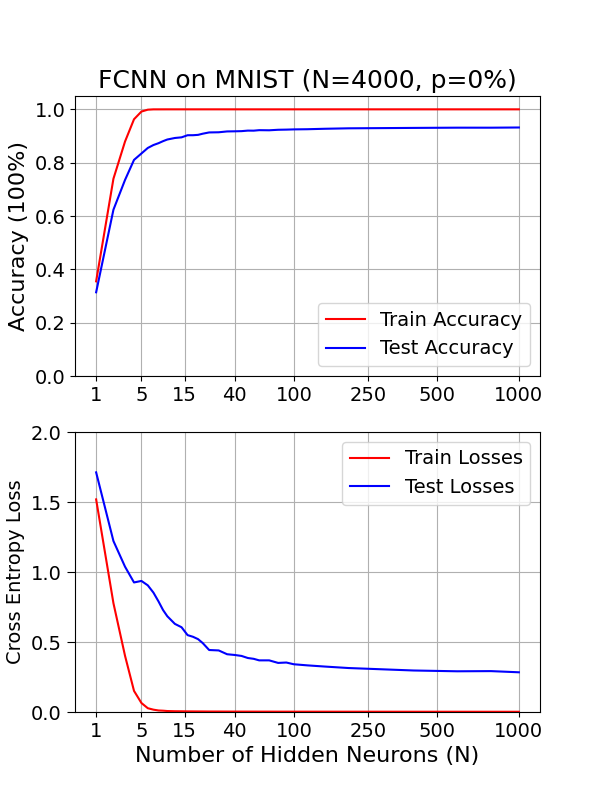}
    \end{minipage} &
    \begin{minipage}[b]{0.3\textwidth}
        \includegraphics[width=\textwidth]{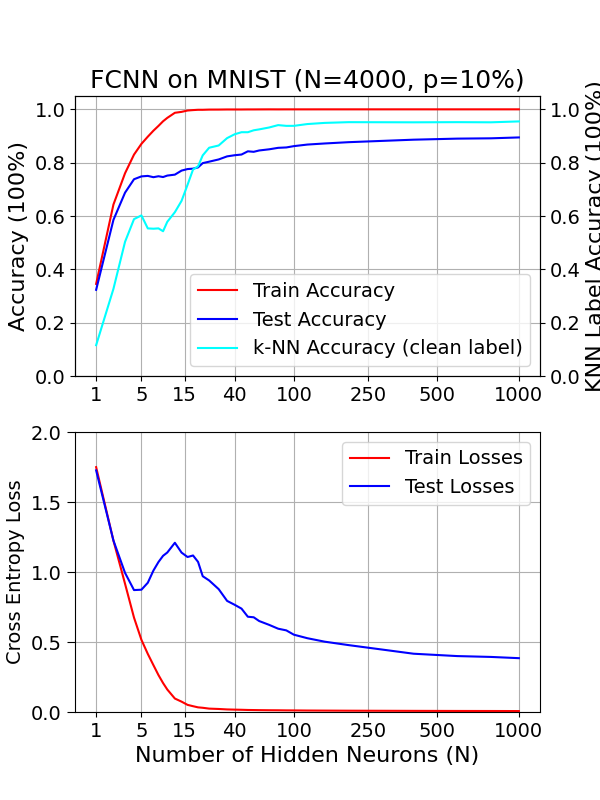}
    \end{minipage} &
    \begin{minipage}[b]{0.3\textwidth}
        \includegraphics[width=\textwidth]{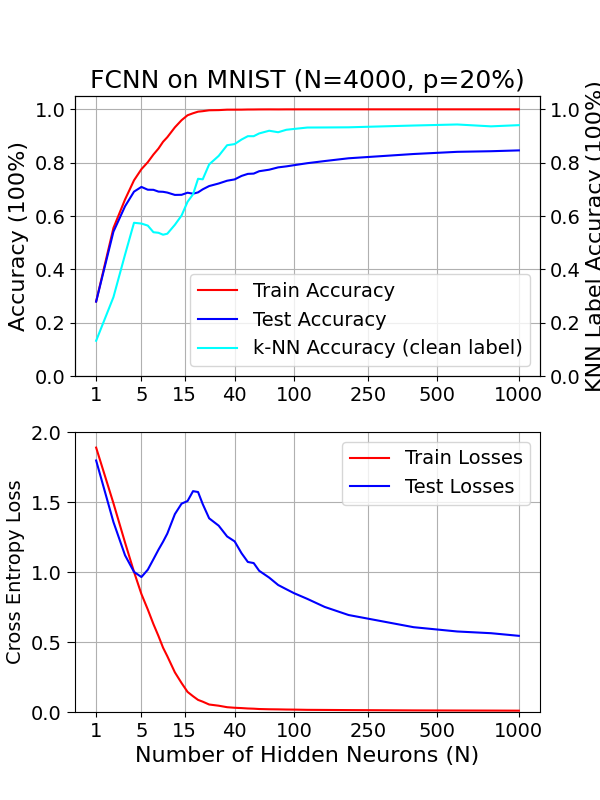}
    \end{minipage} \\
`   \fontsize{8}{8}\selectfont (a) FCNN on MNIST $(p=0\%)$ & 
    \fontsize{8}{8}\selectfont (b) FCNN on MNIST $(p=10\%)$ & 
    \fontsize{8}{8}\selectfont (c) FCNN on MNIST $(p=20\%)$ \\
\end{tabular}
\end{center}
\caption{\label{fig:k-NN-MNIST-FCNN}
    The phenomenon of double descent on two-layer FCNNs trained on MNIST $(N=4000)$, under varying explicit label noise ratios of $p = [0\%, 10\%, 20\%]$ and the prediction accuracy of noisy labelled data denoted as $P$ when $p > 0$. The test error curve of $p = [10\%, 20\%]$ performs the double descent phenomenon and the prediction accuracy $P$ of in-context noisy data learning showed a correlation generalization performance beyond context. 
}
\end{figure}

To comprehend the influence of noisy labelled data on the variation of the learned model during the double descent phenomenon, we examine the percentage $P$ of noisy data, and the majority of its nearest neighbours in feature space are in the same class. In this experiment setup, as depicted in Figure.\ref{fig:k-NN-MNIST-FCNN}(b, c), the trajectory of the $P$ curve follows the pattern observed in the test accuracy curve, indicating a consistent alignment between the two. When test loss first decreases as the model expands, test accuracy and $P$ both rise. With the introduction of label noise, test accuracy decreases while $P$ fluctuates around 70\%. When passing the interpolation threshold, both the test accuracy and $P$ significantly rise to 85\% and nearly 100\%, substantially affirming the validity of our hypothesis. In the meantime, the trend of test loss is opposite to both test accuracy and $P$, while changing correspondingly before and after the interpolation threshold. The computation of percentage $P$ relies solely on the training dataset and does not incorporate unseen data. Consequently, this percentage $P$ may be regarded as a weak predictor of generalization performance, though it necessitates knowledge of the distribution between clean and noisy data.

Recall that the noisy data points will be predicted based on their randomly assigned labels during this stage as the model fully interpolates the training set, we interpret the ascending $P$ curve in the over-parameterized region as a valid `isolation' of the noisy labelled data in the learned feature space. When the neural network positions the learned representation of noisy data points closer to the class associated with their `incorrect' labels, it results in lower $P$ prediction accuracy. Under these circumstances, similar test samples also tend to be predicted incorrectly due to the influence of these noisy labels. If the neural network successfully identifies the underlying data patterns and interpolates the correct samples around noisy data points, the prediction accuracy $P$ will be high by nearby class members. Consequently, test samples resembling the noisy training data have a higher likelihood of being accurately identified due to their proximity to the interpolated surrounding clean data points and becoming less susceptible to the influence of the `isolated' noisy labelled data. This explanation accounts for the improved generalization performance for over-parameterized models.

\begin{figure}[ht]
\begin{center}
\begin{tabular}{ccc}
    \begin{minipage}[b]{0.3\textwidth}
        \includegraphics[width=\textwidth]{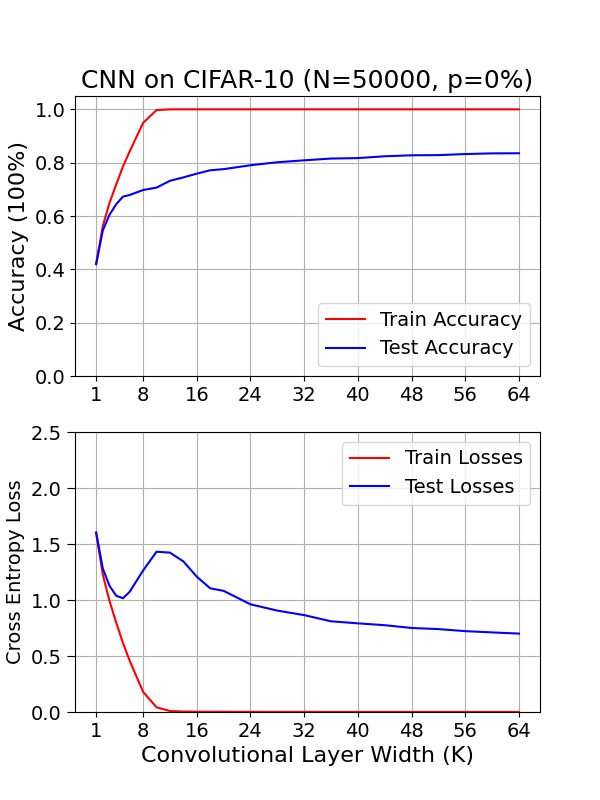}
    \end{minipage}
    &
    \begin{minipage}[b]{0.3\textwidth}
        \includegraphics[width=\textwidth]{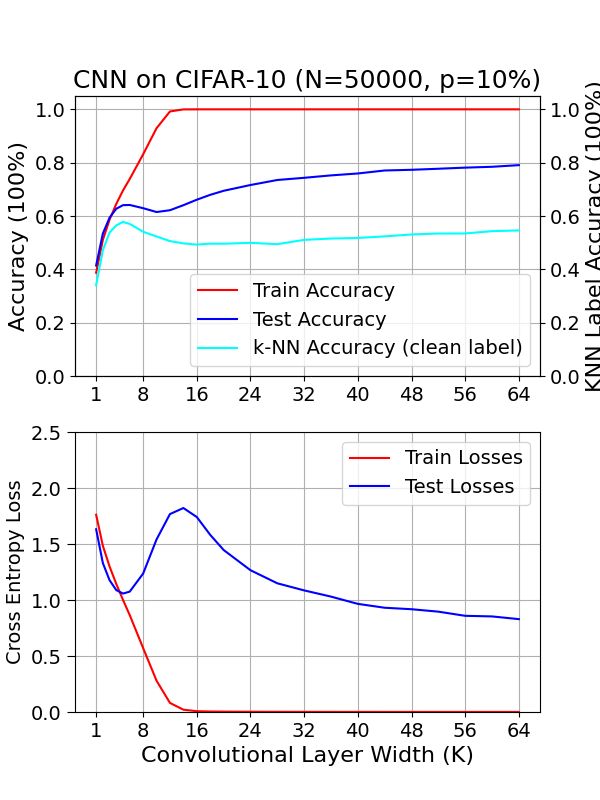}
    \end{minipage} 
    &
    \begin{minipage}[b]{0.3\textwidth}
        \includegraphics[width=\textwidth]{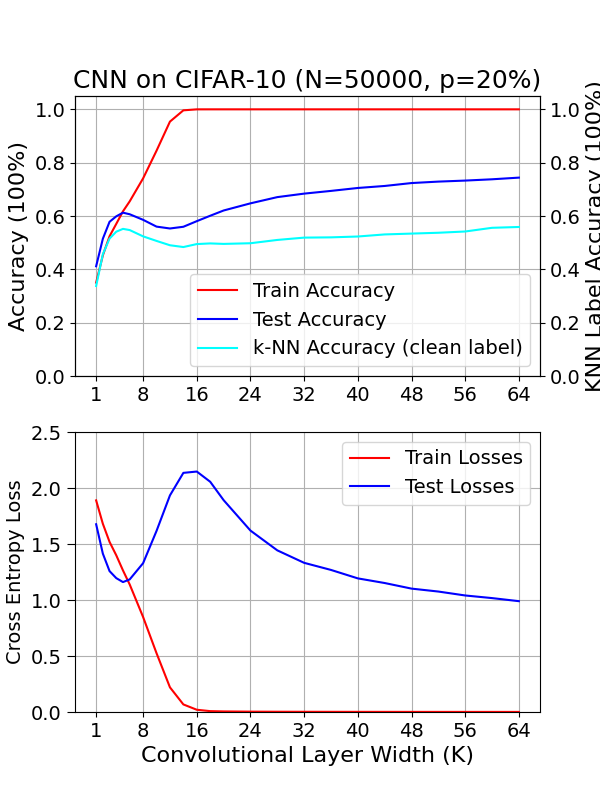}
    \end{minipage} \\
    \fontsize{8}{8}\selectfont (a) CNN on CIFAR-10 $(p=0\%)$ & 
    \fontsize{8}{8}\selectfont (b) CNN on CIFAR-10 $(p=10\%)$ & 
    \fontsize{8}{8}\selectfont (c) CNN on CIFAR-10 $(p=20\%)$ \\
\end{tabular}
\end{center}
\caption{\label{fig:k-NN-CIFAR-10-CNN}
    The phenomenon of double descent on five-layer CNNs trained on CIFAR-10 ($N=50000$), under varying label noise ratios of $p = [0\%, 10\%, 20\%]$ and the prediction accuracy of noisy labelled data denoted as $P$ when $p > 0$. The test error curve of $p = [0\%, 10\%, 20\%]$ performs the double descent phenomenon and the prediction accuracy $P$ of in-context noisy data learning showed a correlation generalization performance beyond context. 
}
\end{figure}

After a baseline case of learning with FCNNs on MNIST, we extended the experiment results to image classification on the CIFAR-10 dataset, using two famous neural network architectures: five-layer CNN and ResNet18. We first visit the experiment when CIFAR-10 ($N=50000$) with zero explicit label noise ($p=0\%$) is trained with CNNs: Figure \ref{fig:k-NN-CIFAR-10-CNN}(a) demonstrated a proposed double descent peak arises around the interpolation threshold. We posit that the variance in observations could stem from a higher ratio of label errors in CIFAR-10 compared to MNIST, as suggested by research estimates~\citep{zhang2017method, northcutt2021pervasive}. When additional label noise ($p=10\%/20\%$) is introduced in Figure \ref{fig:k-NN-CIFAR-10-CNN}(b,c), the test error peak increases accordingly, proving the positive correlation between noise and double descent. While label noise increases the difficulties of models fitting the train set and shifts the interpolation threshold rightwards, the test error peak also shifts correspondingly. 

When we review the noisy labelled data prediction test in the learned feature space, we can also observe an aligned correlation between the trend of generalization performance and prediction accuracy $P$. When test accuracy first increases and then decreases as the model expands, $P$ rises to about 60\% and then decreases before the threshold when the peak of test error arises. Then, as the test error reduces and test accuracy rises, the $P$ curve rises again to about 60\% for over-parameterized models. The observed correlation between the trend of test loss and $P$ remains consistent. An important observation is that the prediction accuracy $P$ demonstrated in the experiments of CNNs is lower compared to those on MNIST. We attribute this variance to the intricate feature space representations, which are more challenging to capture using cosine similarities. Our analysis focused more on how the interpolation strategy varied across the same neural architecture of different widths, and the proposed k-NN methodology only served as an interpretation technique. 

\begin{figure}[ht]
\begin{center}
\begin{tabular}{ccc}
\begin{minipage}[b]{0.3\textwidth}
        \includegraphics[width=\textwidth]{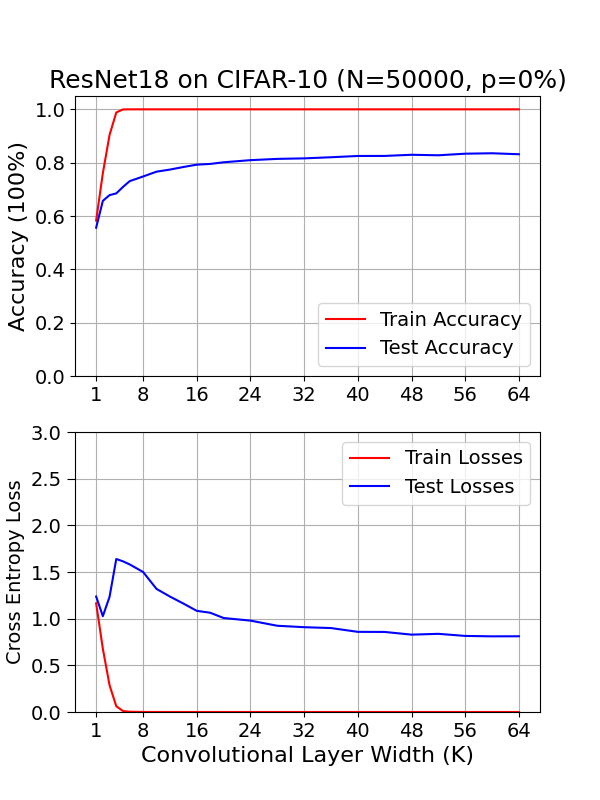}
    \end{minipage}
    &
    \begin{minipage}[b]{0.3\textwidth}
        \includegraphics[width=\textwidth]{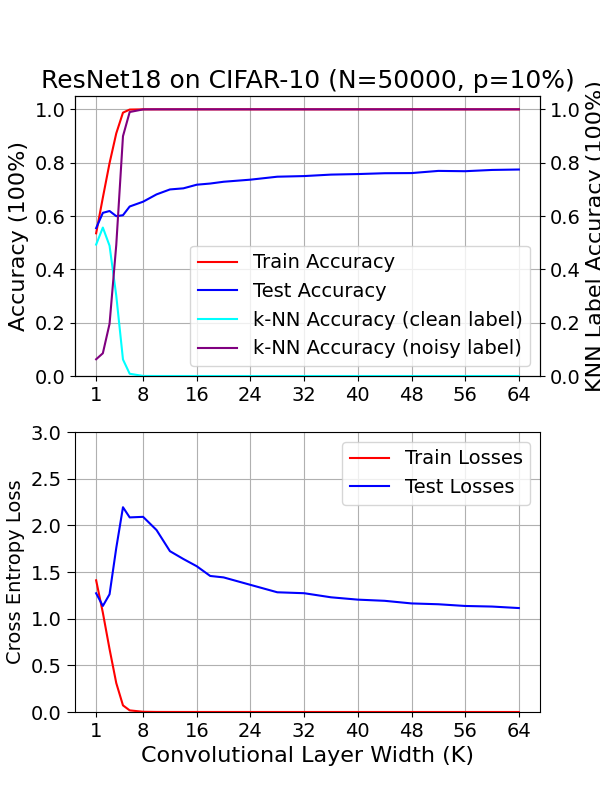}
    \end{minipage} 
    &
    \begin{minipage}[b]{0.3\textwidth}
        \includegraphics[width=\textwidth]{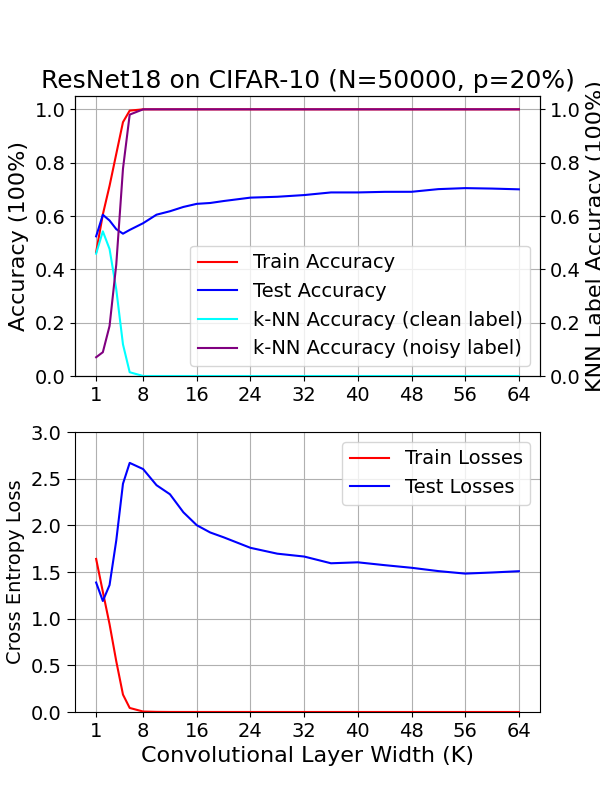}
    \end{minipage} \\
    \fontsize{7}{8}\selectfont (a) ResNet on CIFAR-10 $(p=0\%)$ & 
    \fontsize{7}{8}\selectfont (b) ResNet on CIFAR-10 $(p=10\%)$ & 
    \fontsize{7}{8}\selectfont (c) ResNet on CIFAR-10 $(p=20\%)$ \\
\end{tabular}
\caption{\label{fig:k-NN-CIFAR-10-ResNet18}
    The phenomenon of double descent on ResNet18s trained on CIFAR-10 $(N=50000)$, under varying label noise of $p = [0\%, 10\%, 20\%]$ and the prediction accuracy $P$ when $p > 0$. The test error curve under $p = [0\%, 10\%, 20\%]$ performs the double descent phenomenon. The prediction accuracy $P$ first increases and then decreases to 0 at the interpolation threshold.
}
\end{center}
\end{figure}

However, we may observe in Figure \ref{fig:k-NN-CIFAR-10-ResNet18} of ResNet18s trained and examined on the sample task that w.r.t. increase in model width, $P$ increases before the test error peak, then decreases to zero at the interpolation threshold and remain consistent after since. 
By demonstrating a complementary k-NN prediction accuracy $P^*$ based on the noisy labels rather than the true labels, we observe that for over-parameterized models, the learned representations of the noisy data largely correspond to the incorrect class to which it was assigned. We attribute this observation to the superior feature extraction ability of ResNets compared to CNNs by more convolutional layers and reduced training difficulty by residual connections. 

In this experimental setup, the observation contradicts our hypothesis that over-parameterized models attain a better ability to `isolate' noisy samples. However, it's worth noting that while ResNets outperform CNNs of intermediate scale initially, the introduction of additional label noise leads to over-parameterized CNNs surpassing ResNets by 5\% in test accuracy and even more in test losses. This comparison supports the conclusion that the k-NN prediction capability is linked to the generalization performance of deep learning models concerning their size. Furthermore, this observation leads to a further question: even though the benign over-parameterization phenomenon applies inside various network architectures, the issue of overfitting persists for models with increased expressive capacity.

\section{Summary}

In conclusion, we have explored the feature space of learned representations projected by various neural networks and their interpolation strategies in this chapter. By constructing $P$ to represent the distribution of noisy labelled training data in the learned feature space, we proposed that this k-NN prediction ability of the model is connected to the generalization ability of neural networks. In FCNN trained on MNIST, we observe that as the model size increases, there is clear and distinctive isolation of noisy data points within their original `class-mates' in the learned feature space. We suggest this can be explained as models consistently interpolating correct samples around the noise labels, effectively isolating the misleading effect of the latter. We posit that the mechanism behind this phenomenon can be attributed to two key factors. First, the gradient descent optimization algorithm endeavours to strike a perfect balance among training data points to minimize the loss function, which can lead to the isolation of noisy data points \cite{poggio2017theory}. Second, the characteristics of high-dimensional space may facilitate a beneficial separation of noisy data points from clean data in the learned feature space. 

However, for CNNs and ResNet18s trained on CIFAR-10, we can see as the model's expressive capacity increases (recall discussion from Chapter \ref{chapter4:complexity}), this correlation weakened correspondingly. We attribute this to the model's enhanced capability of extracting features, where more complex data structures, including noise information, can be learned along with the acquisition of more abstract data representations. Through these model comparisons, we can further deduce that while benign over-parameterization exists w.r.t effective model complexity within the same model \citep{nakkiran2021deep}, harmful overfitting persists regarding expressive model capacity across different models. We believe this represents a more comprehensive understanding, which has received less emphasis in previous research. 

Recall the previous findings, we can conclude that for over-parameterized networks, the extended dimensions have led to class-wise activation patterns that correlate less and share less complexity. We have also reviewed the interpolation of noisy samples and the role played by model capacity and model complexity from different perspectives. It is believed that these reviewed characteristics lead to better generalization performance and the `secret' mechanism against overfitting noise and unrepresentative data. Additional discussions will be presented in the conclusion chapter of this study, along with supplementary experiment results provided in the appendix.

\chapter{Conclusions\label{chapter6:conclusion}}

This paper reproduced the double descent (benign over-parameterization / benign over-fitting) phenomenon across various deep learning architectures and explored various hypotheses emphasizing the interplay between class-wise activations and interpolation of samples in the hidden feature space from an empirical perspective. While the empirical validation may not be extended to further neural architectures (as served by theoretical analysis), this paper aims to provide insights into the intrinsic mechanisms of why big deep learning models generalized so well without the involvement of regularization techniques and bridge the gap between theoretical studies and empirical practises. Furthermore, we also targeted re-framing the current question scope, thereby facilitating further explorations to solutions unravelling the `mysteries' of deep learning.

\section{Evaluation}

The three major hypotheses explored in this paper are 1. neural networks exhibit class-wise activation patterns that are more distinctive in over-parameterized regimes (Chapter \ref{chapter3:activation}); 2. the richness of these class-wise activation patterns served as an indication of model complexity, further related to the generalization performance (Chapter \ref{chapter4:complexity}); 3. over-parameterized neural networks can `isolate' noisy samples in the hidden feature space, thus obtaining superior generalization ability (Chapter \ref{chapter5:interpolation}). We'll consolidate the investigations of these hypotheses and assess the accomplishments in each chapter individually.

For the first hypothesis (activation patterns), we have defined Class Activation Matrices (CAMs) and presented the similarity heatmap and mean similarities curve w.r.t. increase in model width in Chapter \ref{chapter3:activation}. We have demonstrated the similarities between CAM contains interpretable features between classes, representing the learning ability of neural networks. We further showcased that the representation layer of various neural architectures exhibits a higher similarity across class-wise activations than the classification layer. Furthermore, these class-wise activation similarity tends to increase before the interpolation threshold and decrease after on the representation layer, while it decrease to zero in the over-parameterized regime for the classifier. The observations validate our hypothesis to some extent, though does not provide sufficient explanation for the descent in the over-parameterized regime. On the other hand, it remains doubtful if cosine similarity served as the best method to access matrix similarities. The Pearson correlation coefficient we examined has produced similar results, while further methods are left to be explored.

Next, speaking of the second hypothesis (richness/complexity), we have proposed an empirical technique for estimating the richness of a given set (function), which is further referred to as the complexity of accessed class-wise activation patterns in Chapter \ref{chapter4:complexity}. The exploration in this chapter continues from the last, and the proposed experiment results on these intrinsic model complexities demonstrated a surprisingly good alignment with the double descent phenomenon in most of the test cases. We believe that by combining the results of class-wise activation's independence and estimated complexity, rich insights have been provided into classification neural networks on how over-parameterization prevents over-fitting and promotes lower variance and improved generalization performance. It remains arguable on the correctness and robustness of the proposed methodology. We believe the robustness is guaranteed by extensive experiments on different parameters, while its relationship to the true model complexity is not clear. Nevertheless, the strong alignment between the two is still a surprising and intriguing phenomenon to explore.

Finally, in Chapter \ref{chapter5:interpolation} we have explored the geometry of the class-wise activations: as data representations in the learned feature space, focusing on the interplay of clean and noisy samples, when external label noise is introduced. By demonstrating experiment results through the proposed k-NN prediction methodology, we demonstrated that our hypothesis is robust over FCNNs, less significant in CNNs and oppositely observed in ResNets. These could be attributed to the increasing learning ability across the studied neural architectures, that the injection of input features becomes more abstract in the hidden feature space, and more complex class feature patterns can be learned, even when noise is introduced. Thus we may see that with our hypothesis violated in ResNets, the corresponding test accuracy on CIFAR-10 is damaged in the over-parameterized regime compared to CNNs of the same scale, despite its superior ability in feature extraction. This experiment proves that while over-parameterization promotes generalization in the same model architecture, increased model complexity across different models may still suffer from over-fitting issues, which is a warning to future deep learning practises. 

Besides the three hypotheses discussed in the main context of this paper, some additional questions regarding the implicit sparsity of both weight and activations, high label noise scenarios and choice between different neural architectures are also included in this study with results and discussions presented in the Appendix. In the next and last section, we will look into the limitations of this study and discuss some future directions towards the generalization theory of modern deep learning techniques.

\section{Future Works}

This study also highlights certain potential future directions, discerned through an analysis of its limitations and weaknesses. In this section, we aim to explore some potential avenues that could prove beneficial to the field. 

Firstly, the empirical studies of this study could be extended to further experiment setups. Transformers, one of the most popular neural architectures extensively studied and applied nowadays could be included, including the studied architectures with different parameters and architectural choices. Different optimizers (hyperparameter settings) and training schemes are an intriguing direction to explore as well. While this study focused on the class-wise activations in neural networks trained for classification tasks, the question remains that if the proposed hypothesis can be re-framed for regression tasks, and if some experiments can be conducted on more general-purpose large models (LLMs, Multi-models) to bridge into comprehensive analysis of their properties (e.g. the scaling law). 

Next, let us overview the potential future works on analyzing the role of various `hidden factors' in shaping the generalization performance of deep learning models. Based on the study of class-wise activation patterns, similar methodologies may promote the study of interpretable machine learning. Explicit regularization techniques for constraining these hidden patterns may further benefit generalization and transfer learning to new tasks by introducing new patterns from a limited set of samples. We have also analyzed the interpolation of training samples in the hidden feature space in this paper which may benefit from perspectives of high-dimensional statistics. Exploring the properties of feature distribution in space of different dimensions and analyzing the volume contraction is a promising direction to explore. Speaking of the discussion of the feature extraction process employed by different neural models presented in this paper, that could be characterized by the information theory on feature injections through layers. Certain hidden factors or perspectives are not discussed in this paper, including implicit sparsity (roughly discussed in Appendix \ref{appendix:sparsity}), sharpness (smoothness) of interpolation \citep{gamba2022deep} and regularization techniques, in which we all see potential avenues for exploration in the future.

Lastly, the most significant limitation of this study is its emphasis on research solely from empirical perspectives: hypothesis formation and validations. This approach lacks essential causal analysis of the `double descent; phenomenon, limiting the applicability of its conclusions to neural networks under varied experimental setups. Additional experiments would be required to address this gap. In traditional statistical theory and practises, the bias-variance trade-off (decomposition) has been a universal understanding of the property of machine learning algorithms, while it is now proven as a limited understanding by the existence of the double descent phenomenon. It is expected that a comprehensive theoretical framework could be `re-established' for general deep learning techniques, proposing the role of different factors (architectural choices, training schemes, tasks and datasets) and offering a complete understanding of this intriguing phenomenon. The ultimate goal of research on double descent is expected to provide design principles for deep learning models, promoting efficiency in building proficient and robust software systems for solving real-world problems.

\bibliographystyle{ACM-Reference-Format}
\addcontentsline{toc}{chapter}{Bibliography}
\bibliography{refs}

\appendix
\addcontentsline{toc}{chapter}{Appendix\label{appendix}}
 
\chapter{Scaling of Model Parameterization\label{appendix:scaling}}
\begin{figure}[ht]
\begin{center}
\begin{tabular}{ccc}
    \begin{minipage}[b]{0.3\textwidth}
        \centering
        \includegraphics[width=\textwidth]{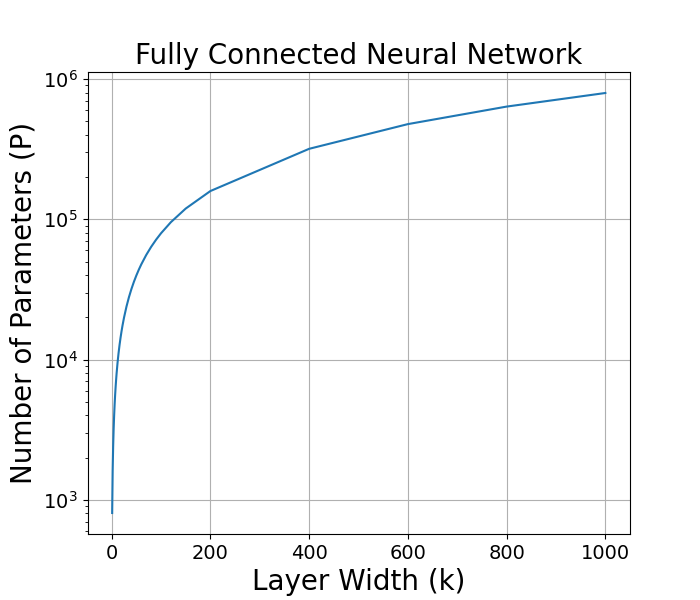}
        \label{fig:FCNN}
    \end{minipage}
    &
    \begin{minipage}[b]{0.3\textwidth}
        \centering
        \includegraphics[width=\textwidth]{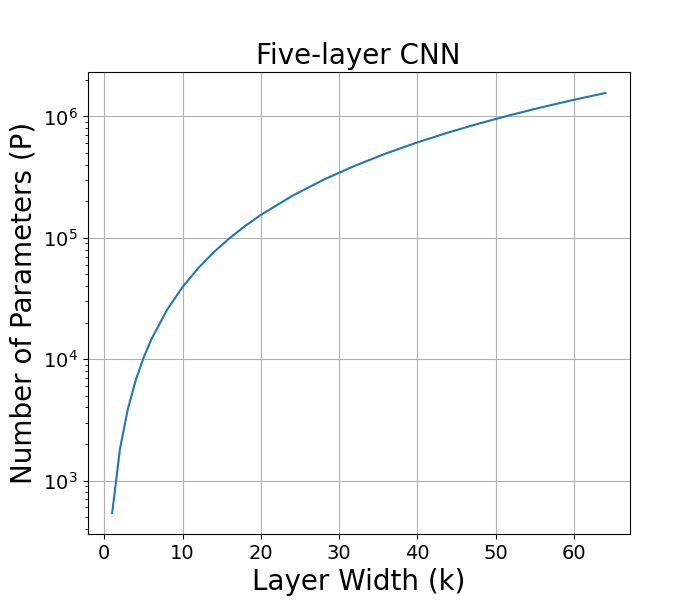}
        \label{fig:CNN}
    \end{minipage} 
    &
    \begin{minipage}[b]{0.3\textwidth}
        \centering
        \includegraphics[width=\textwidth]{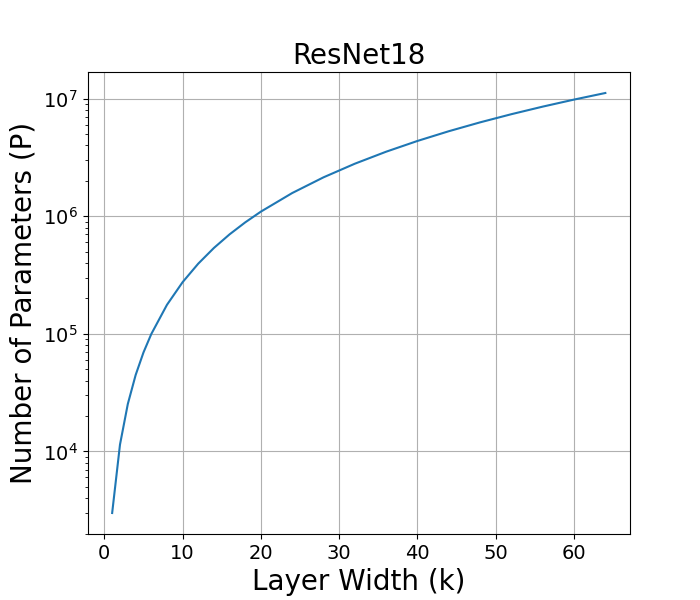}
        \label{fig:ResNet18}
    \end{minipage} \\
    (a) Two-layer FCNNs & 
    (b) Five-layer CNNs & 
    (c) ResNet18s \\
\end{tabular}
\end{center}
\caption{\label{fig:Scaling of Parameterization}
    Scaling of the number of parameters $P$ as model size with layer width unit $k$ of the three neural architectures: two-layer fully connected neural networks, five-layer CNNs and ResNet18s used in our experiments. A logarithmic scale is applied to all neural architectures' parameter counts $P$.
}
\end{figure}

The scaling of model parameterization to the universal layer width unit $K$ is shown in Figure.\ref{fig:Scaling of Parameterization}. We may see that the scaling of FCNNs and CNNs are similar with differences further neglectable with an additional scaling factor of $(1/4, 4)$ adopted to the FCNN when presenting the experiment results. The overall scaling of ResNet18s is higher than the five-layer CNNs studies due to increased layer depth on each model width $k$. All model-wise experiment results presented in this paper are x-axis labelled as this model width parameter $k$ instead of scaling the number of parameters $P$. Nevertheless, we believe this labelling does not impact out conclusion due to the monotonic relationship between these two factors.

\chapter{Discussion on Implicit Activation Sparsity in Neural Networks\label{appendix:sparsity}}
Sparsity has a long history of exploration in deep learning, with pruning emerging as a prevalent regularization technique aimed at mitigating overfitting \cite{reed1993pruning, wen2016learning}. There's even evidence of a sparsity-wise double descent phenomenon, where increasing model sparsity through pruning initially overfits, then alleviates, and overfits again as valuable information may be ultimately discarded \cite{he2022sparse}. In contrast to research investigating the effects of pruning on over-parameterized models \citep{chang2021provable}, we underscore that our experimental setups do not employ explicit sparsification techniques. Therefore no explicit sparsification occurs, rendering the relevant previous discussions inapplicable. Thus, we attempted to analyse the implicit sparsity attained in model architecture trained with SGD or in feed-forward activations. If implicit sparsity indeed manifests in neural models without sparsification methodologies, we further endeavoured to comprehend its potential role in the double descent phenomenon, and whether it mirrors the function of explicit sparsity.

\begin{figure}[h]
    \centering
    \includegraphics[width=0.6\textwidth]{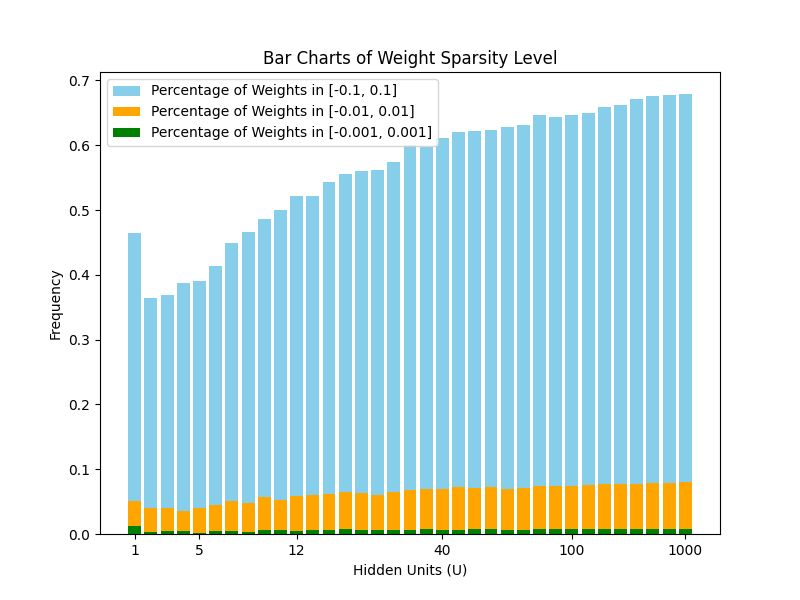}
    \caption{The mean ratio of sparse neurons with a weight near zero of different degrees of two-layer FCNNs trained on MNIST with no label noise.}
    \label{fig:Weight-Sparsity}
\end{figure}

We first study weight sparsity in fully connected neural networks, which is shown to be non-significant in Figure.\ref{fig:Weight-Sparsity}. The figure presented the ratio of sparse neurons with a weight near zero of different degrees, from which we may conclude that the weight with an absolute value of less than 0.1 increases w.r.t. increase in neuron numbers. This observation may correspond to the hypothesis outlined in \cite{teague2022geometric}. Nevertheless, we can see $< 5\%$ of neurons are in the range of $[-0.01, 0.01]$ and no weight penalized to zero. We may conclude from this observation that sparsity in model architecture is not a major contributor to the benign over-parameterization phenomenon reported in previous studies and Chapter \ref{chapter2:experiments}.

In addition to sparsity in architecture, implicit sparsity is observed during the feed-forward process of neural networks, specifically in the activation of the model. In addition to sparsity in model architecture, \citeauthor{muthukumar2023sparsity} concluded that deep Feed Forward ReLU networks take advantage of this sparsity that is achieved in the hidden layer activations \citep{muthukumar2023sparsity}. A hypothesis has thus been raised that over-parameterized models utilized this activation sparsity to promote better performance than under-parameterized models. In this study, we have also conducted a brief empirical study on the activation sparsity presented in the experiment set-ups, with experiment results shown in the next section. 

\section{Activation Sparsity}

\begin{figure}[ht]
\begin{center}
\begin{tabular}{ccc}
    \begin{minipage}[b]{0.3\textwidth}
        \includegraphics[width=\textwidth]{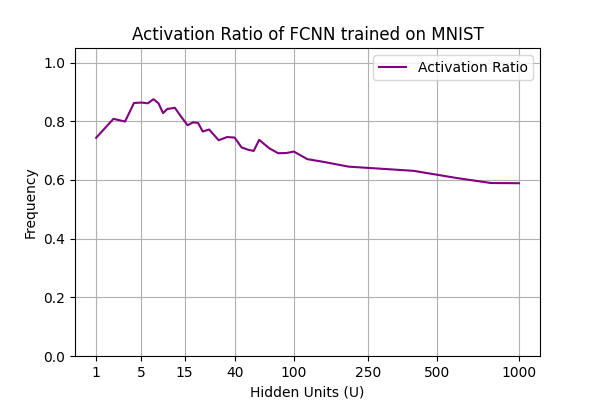}
    \end{minipage} &
    \begin{minipage}[b]{0.3\textwidth}
        \includegraphics[width=\textwidth]{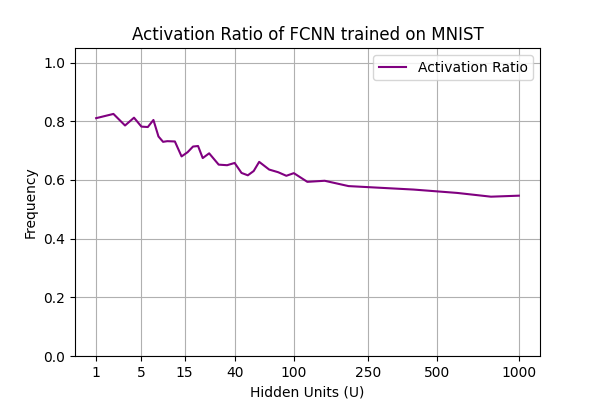}
    \end{minipage} &
    \begin{minipage}[b]{0.3\textwidth}
        \includegraphics[width=\textwidth]{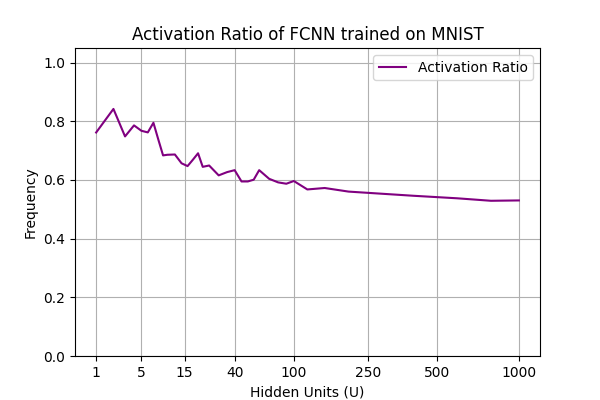}
    \end{minipage} \\
    \fontsize{8}{8}\selectfont 1.(a) SimpleFC on MNIST $(p=0\%)$ &
    \fontsize{8}{8}\selectfont 1.(b) SimpleFC on MNIST $(p=10\%)$ &
    \fontsize{8}{8}\selectfont 1.(c) SimpleFC on MNIST $(p=20\%)$ \\
\end{tabular}
\end{center}
\caption{\label{fig:Activ-Ratio-FCNN}
    The mean activation ratio of hidden neurons of each test image of two-layer FCNNs trained on MNIST.
}
\end{figure}

We measure activation sparsity simply as the mean ratio of non-zero activations of hidden neurons on each test data point. By analysing Figure \ref{fig:Activ-Ratio-FCNN}, we may observe that the mean ratio of non-zero activations over data points decreases w.r.t. increase of model width $k$. Nonetheless, the specific trend of sparsity varies across various label noise ratios: In the absence of label noise, sparsity initially increases for $k < 8$ and then decreases steadily thereafter. Conversely, when label noise is introduced, sparsity initially fluctuates and continues to decrease further with an increase in label noise. The induced sparsity by introducing label noise is believed to be an intriguing phenomenon to be explored. Though the implicit sparsity driven by model width comes with strong evidence in this and previous studies, a theoretical or casual relationship to the benign over-parameterization phenomenon is rather unclear.

\begin{figure}[ht]
\begin{center}
\begin{tabular}{ccc}
    \begin{minipage}[b]{0.3\textwidth}
        \includegraphics[width=\textwidth]{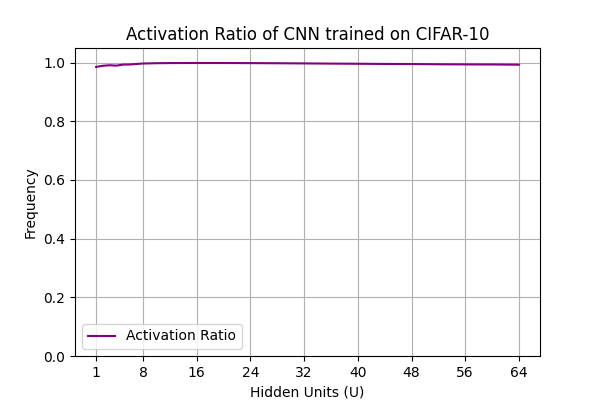}
    \end{minipage} &
    \begin{minipage}[b]{0.3\textwidth}
        \includegraphics[width=\textwidth]{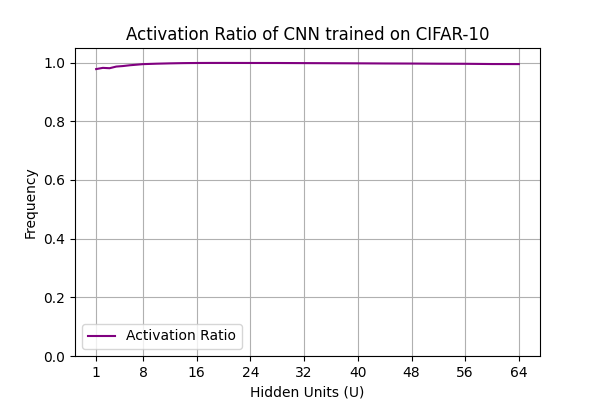}
    \end{minipage} &
    \begin{minipage}[b]{0.3\textwidth}
        \includegraphics[width=\textwidth]{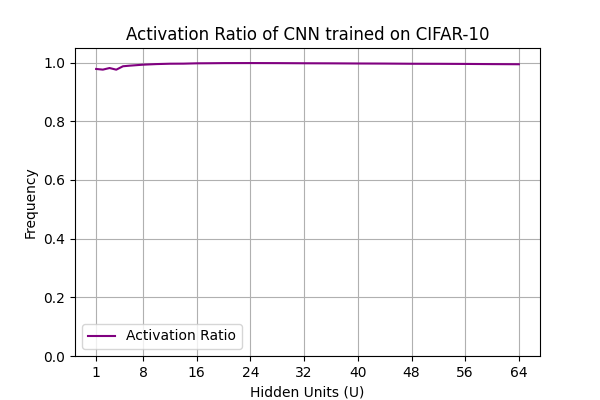}
    \end{minipage} \\
    \fontsize{8}{8}\selectfont (a) CNN on CIFAR-10 $(p=0\%)$ &
    \fontsize{8}{8}\selectfont (b) CNN on CIFAR-10 $(p=10\%)$ &
    \fontsize{8}{8}\selectfont (c) CNN on CIFAR-10 $(p=20\%)$ \\
\end{tabular}
\end{center}
\caption{\label{fig:Activ-Ratio-CNN}
    The mean activation ratio of hidden neurons of test images on CIFAR-10 trained on CNN.
}
\end{figure}

Conversely, with convolutional layers for efficient feature extraction and the relatively small dimension in model width, five-layer CNNs and ResNet18s do not demonstrate implicit sparsity in their feed-forward activations. This further proves that implicit sparsity is not the main cause of the benign over-parameterization phenomenon in deep learning practices. On the other hand, it also proves that the class activation pattern differentiation discussed in Chapter 3 does not always relate to turned-off activations. One further experiment is carried out on the FCNNs to unravel when implicit sparsity is induced if the neuron's activation has a more specified choice of class features. Thus, we proposed to compute the Normalized Discounted Cumulative Gain on each neuron's activation frequency of each class, which is designed to reflect this specification.

\section{Class-wise Neural Activation Specification}

With the growth of model width and the accompanying increase in activation sparsity, a valid hypothesis is that the class functions are `disaggregated' on neuron deciders. This couples the decrease in function complexity and potentially an explanation of the benign over-parameterization phenomenon. The experiment design here referenced the idea of Normalized Discounted Cumulative Gain (NDCG) in the evaluation of Information Retrieval. Each neuron will gain more on activations of the class with more frequent self-activations (which means it is likely to be a decider of that class) and gain less the other way around. The cumulative gains are averaged and normalized on a scale of $[0-1]$. The neural NDCG on class prediction frequency attains its minimal at 0 when all neurons are activated equally on all class predictions and attains its maximum at 1 when all neurons are only activated on one specific class.

\begin{figure}[ht]
\begin{center}
\begin{tabular}{ccc}
    \begin{minipage}[b]{0.3\textwidth}
        \includegraphics[width=\textwidth]{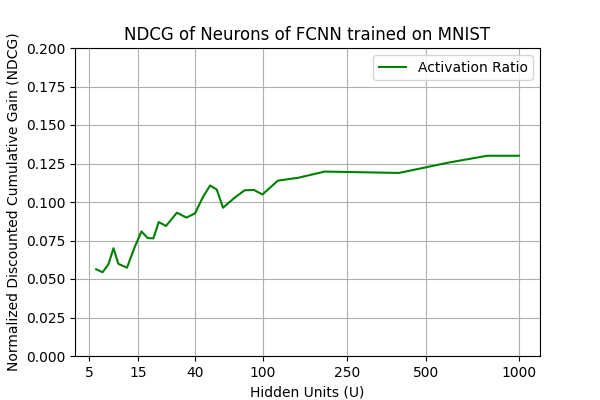}
    \end{minipage} &
    \begin{minipage}[b]{0.3\textwidth}
        \includegraphics[width=\textwidth]{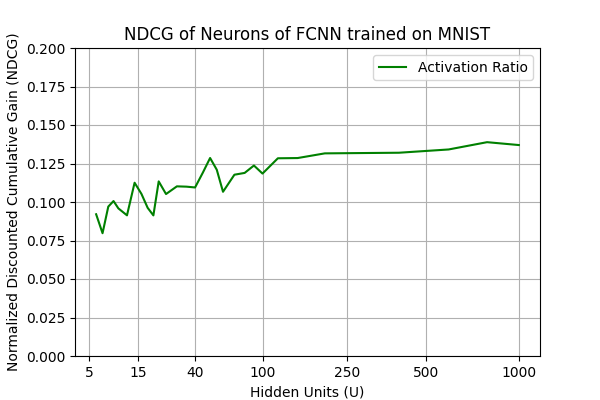}
    \end{minipage} &
    \begin{minipage}[b]{0.3\textwidth}
        \includegraphics[width=\textwidth]{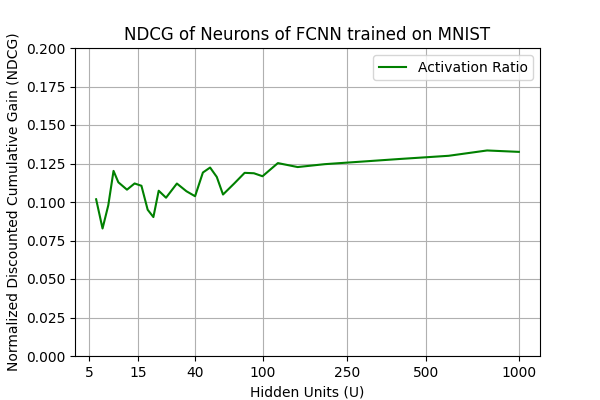}
    \end{minipage} \\
    \fontsize{8}{8}\selectfont (a) FCNN on MNIST $(p=0\%)$ &
    \fontsize{8}{8}\selectfont (b) FCNN on MNIST $(p=10\%)$ &
    \fontsize{8}{8}\selectfont (c) FCNN on MNIST $(p=20\%)$ \\
\end{tabular}
\end{center}
\caption{\label{fig:NDCG-FCNN}
    The Neural Normalized Discounted Cumulative Gain (NDCG) Value on class prediction frequency. The NDCG Values fluctuate over tiny models and share the same tendency to increase with the number of parameters though in a small range.
}
\end{figure}

From the experiment results depicted in Figure \ref{fig:NDCG-FCNN}, we can see that despite an increase in meaningless small models, the overall trend of class sparsity on neuron activation decreases with the increase in model width. Similar to various previous analyses, increased noise ratio slightly induced higher specification on each neuron. This result explains the decrease in class complexity to some extent, though we are comparing the activation specification in this case. On the other hand, the overall increase in NDCG values is not obvious and the numerical values are not significant over the scale of $[0-1]$. Thus, the analysis here can only be utilized as supplement materials to understand the activation of FCNNs, instead of accounting for the mechanism behind the benign over-parameterization phenomenon. 

\chapter{Experiment Results with High Label Noise Ratio\label{appendix:noise}}
In the main context of this paper, while we have explored the introduction of label noise, our focus has primarily been on a relatively small label noise ratio. In this chapter, we will present the same experimental results across chapters 2, 3, 4, and 5 but with higher label noise ratios of $p=[40\%, 60\%]$, comparing them to the previously discussed scenario of $p=20\%$. This comparison aims to demonstrate the impact of noisy data on shaping neural network performance and its intrinsic mechanisms.

\section{Experiment Results and Activation Correlation}

\begin{figure}[h]
\begin{center}
\begin{tabular}{ccc}
    \begin{minipage}[b]{0.3\textwidth}
        \includegraphics[width=\textwidth]{images/Experiments/MNIST-FCNN-Epochs=4000-p=20-U.png}
    \end{minipage} &
    \begin{minipage}[b]{0.3\textwidth}
        \includegraphics[width=\textwidth]{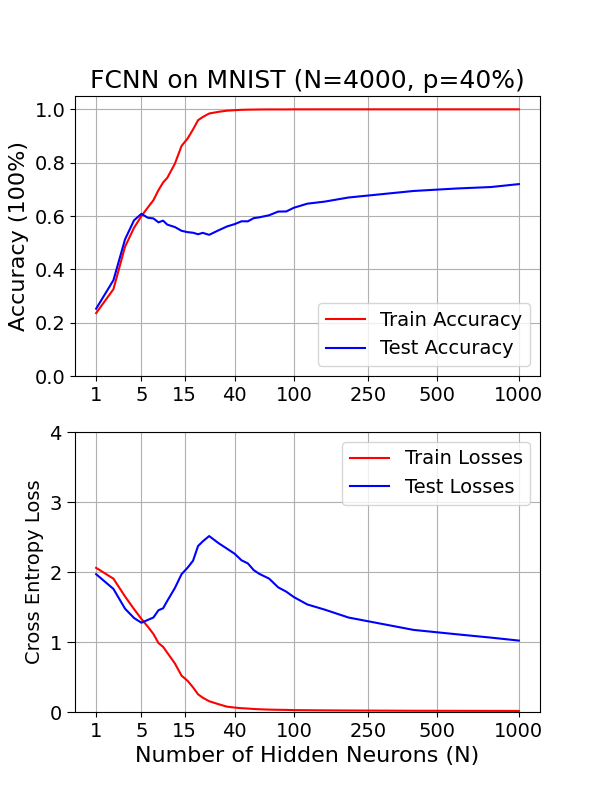}
    \end{minipage} &
    \begin{minipage}[b]{0.3\textwidth}
        \includegraphics[width=\textwidth]{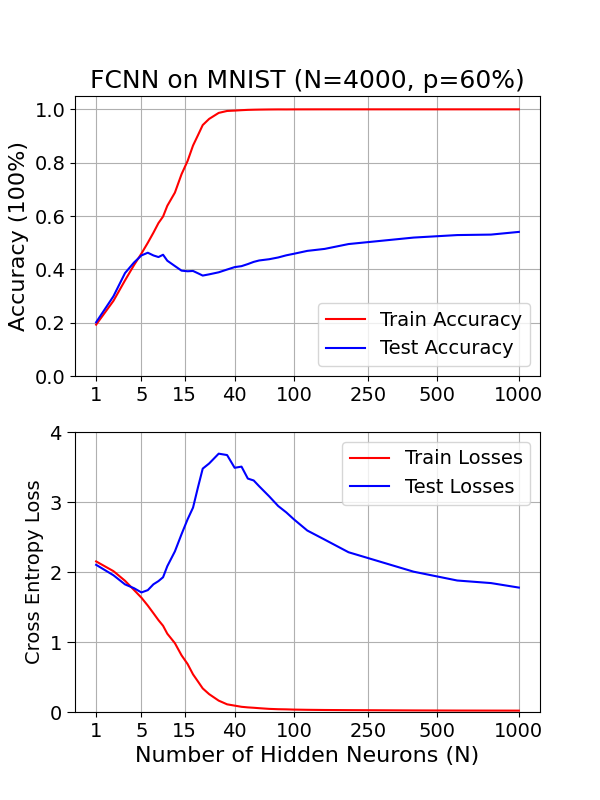}
    \end{minipage} \\
    \begin{minipage}[b]{0.3\textwidth}
        \includegraphics[width=\textwidth]{images/Activation/Act-Corr-MNIST-FCNN-Epochs=4000-p=20.png}
    \end{minipage} &
    \begin{minipage}[b]{0.3\textwidth}
        \includegraphics[width=\textwidth]{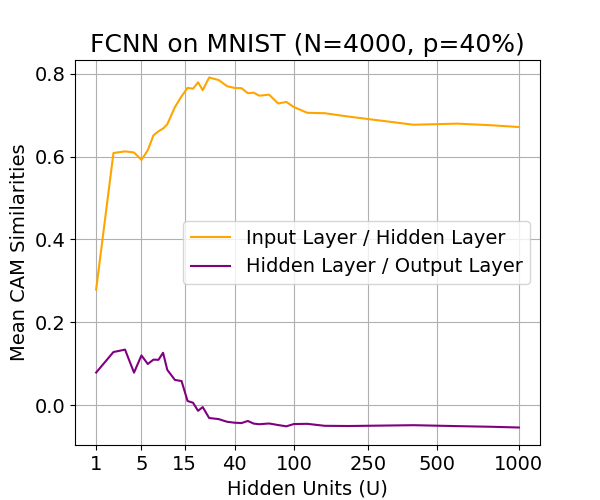}
    \end{minipage} &
    \begin{minipage}[b]{0.3\textwidth}
        \includegraphics[width=\textwidth]{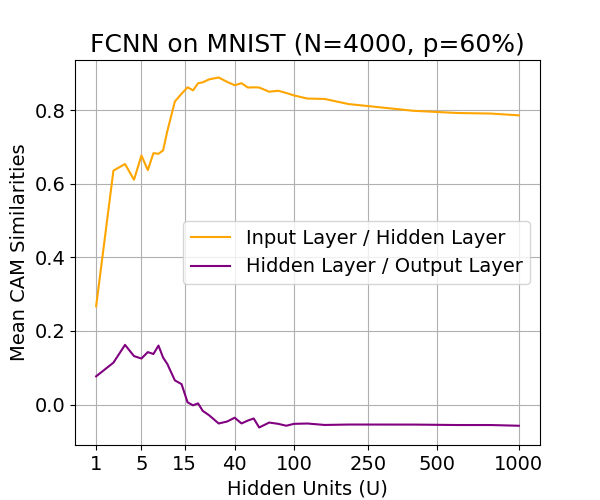}
    \end{minipage} \\
    \fontsize{8}{8}\selectfont (a) FCNN on MNIST $(p=20\%)$ &
    \fontsize{8}{8}\selectfont (b) FCNN on MNIST $(p=40\%)$ &
    \fontsize{8}{8}\selectfont (c) FCNN on MNIST $(p=60\%)$ \\
\end{tabular}
\end{center}
\caption{\label{fig:FCNN-MNIST-Noise}
    The experiment results and mean similarities across all CAMs of two-layer FCNNs trained on MNIST $(N=4000)$ w.r.t. increase in model width $k$ under label noise ratio $p=[20\%, 40\%, 60\%]$. The yellow line denotes the activation similarities between the input layer (layer $f_0$) and the hidden layer (layer $f_1$), whereas the purple line signifies the similarities between the hidden layer (layer $f_1$) and the output layer (layer $f_2$).
}
\end{figure}

Figure \ref{fig:FCNN-MNIST-Noise} illustrates the double descent phenomenon observed in two-layer FCNNs trained on MNIST with a higher label noise ratio ($p=[40\%, 60\%]$) compared to $p=20\%$. It is evident that with increased label noise, the double descent phenomenon becomes more pronounced, resulting in a significant reduction in generalization performance. The interpolation threshold shifts from $k=20$ to $k=40$, accompanied by a roughly 30\% decrease in test accuracy. Another noteworthy observation is that in scenarios where no or a small ratio of label noise is applied to the training set, over-parameterized networks can achieve considerably smaller test errors than under-parameterized ones. However, with $p=60\%$, the test loss at $k=1000$ becomes comparable to the 'sweet-spot' observed at $k=5$.

We then examine the correlation between class-wise activation patterns. We observe a similarity curve with richer activities between the layers under examination. With high label noise, during the underfitting stage, the similarity between both the representation layer and the classifier layer increases. At the `sweet-spot', where a bias-variance trade-off is achieved, both similarities fluctuate within a fixed value range. Next, as the neural network begins to overfit, the similarity of the representation layer increases dramatically while the similarity of the classifier decreases. It is noticeable that increasing the noise ratio leads to an increase in the numerical value of the similarity peak between the representation layer, demonstrating the randomness introduced between labelled images. Finally, upon passing the interpolation threshold, the similarity between the representation layer decreases again, while the classifier remains consistent, exhibiting a similar trend to that observed with low label noise. Further research and explanation are needed to fully understand the dynamics underlying these observations.

\begin{figure}[h]
\begin{center}
\begin{tabular}{ccc}
    \begin{minipage}[b]{0.3\textwidth}
        \includegraphics[width=\textwidth]{images/Experiments/CIFAR-10-CNN-Epochs=200-p=20-U.png}
    \end{minipage} &
    \begin{minipage}[b]{0.3\textwidth}
        \includegraphics[width=\textwidth]{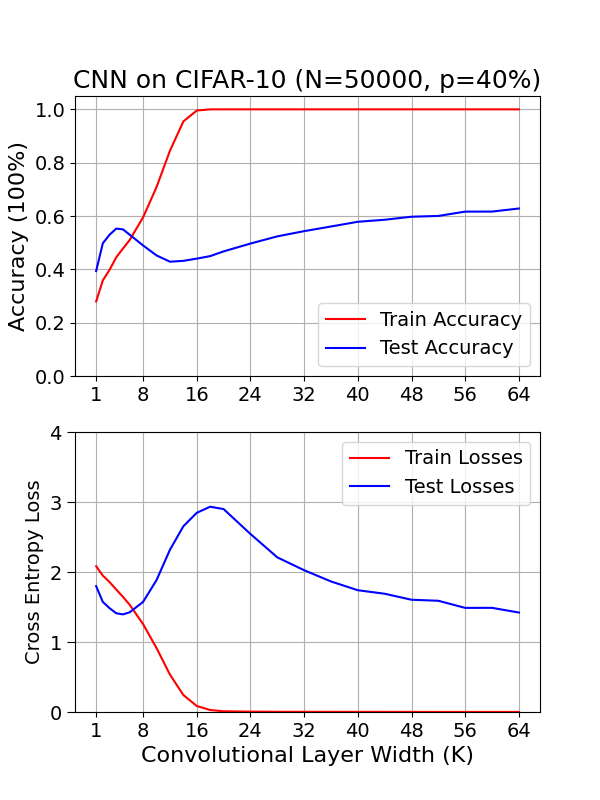}
    \end{minipage} &
    \begin{minipage}[b]{0.3\textwidth}
        \includegraphics[width=\textwidth]{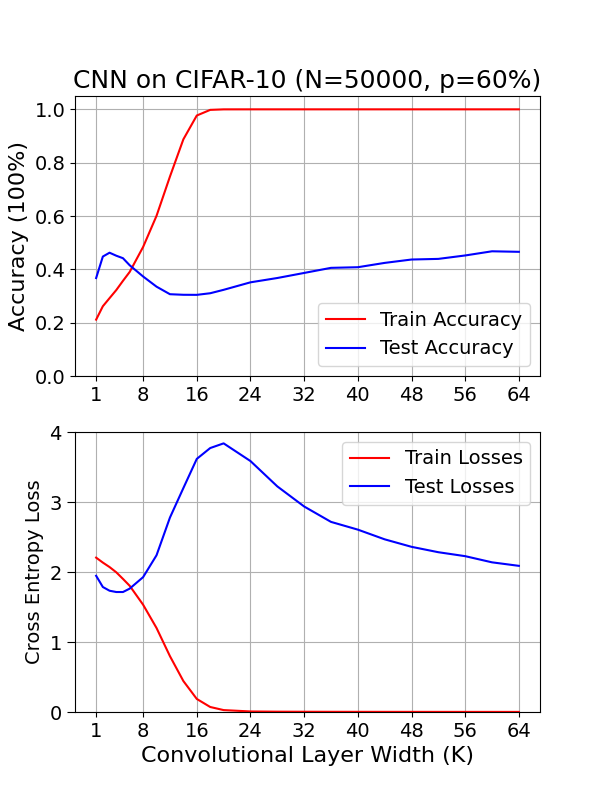}
    \end{minipage} \\
    \begin{minipage}[b]{0.3\textwidth}
        \includegraphics[width=\textwidth]{images/Activation/Act-Corr-CIFAR-10-CNN-Epochs=200-p=20.png}
    \end{minipage} &
    \begin{minipage}[b]{0.3\textwidth}
        \includegraphics[width=\textwidth]{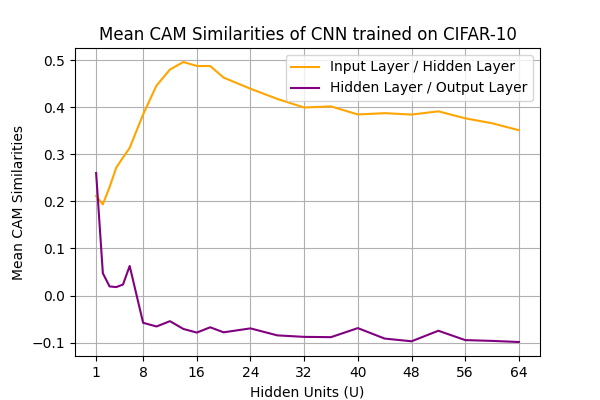}
    \end{minipage} &
    \begin{minipage}[b]{0.3\textwidth}
        \includegraphics[width=\textwidth]{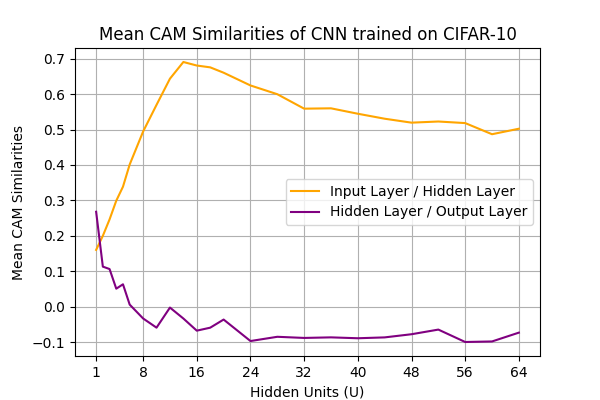}
    \end{minipage} \\
    \fontsize{8}{8}\selectfont (a) CNN on CIFAR-10 $(p=20\%)$ &
    \fontsize{8}{8}\selectfont (b) CNN on CIFAR-10 $(p=40\%)$ &
    \fontsize{8}{8}\selectfont (c) CNN on CIFAR-10 $(p=60\%)$ \\
\end{tabular}
\end{center}
\caption{\label{fig:CNN-CIFAR-10-Noise}
    The experiment results and mean similarities across all CAMs of five-layer CNNs trained on CIFAR-10 $(N=50000)$ w.r.t. increase in model width $k$ under label noise ratio $p=[20\%, 40\%, 60\%]$. The yellow line denotes the activation similarities between the input layer (layer $f_0$) and the hidden layer (layer $f_1$), whereas the purple line signifies the similarities between the hidden layer (layer $f_1$) and the output layer (layer $f_2$).
}
\end{figure}

As similarly demonstrated in FCNN trained on MNIST, when high label noise ($p=[40\%, 60\%]$) is introduced on CIFAR-10 and trained with fiver-layer CNNs, the test error increases correspondingly with a significant drop in test accuracy, even in the over-parameterized regime. The underfitting stage is much shorter while models quickly overfit the additional noise persisted in the training dataset. On the other hand, the interpolation threshold is shifted rightwards, as additional model capacity is required to fully fit the added noise information. When examining the correlation between class-wise activation patterns, we may observe our previous conclusion being much more strongly supported by the line curves: For the representation layer(s), this correlation increases in the under-parameterized regime and decreases in the over-parameterized regime; and for the classifier, this correlation decreases before the interpolation threshold and remain consistent after since. We may even make one unreasonable connection between (class-wise correlation of) the classifier to the training error curve, and of the representation layer(s) to the test error curve. We are pleased to confirm that these observations remain consistent even in test cases with significant label noise, and we are prepared to explore the precise mechanism further in future studies.

\section{Class-wise Activation Complexity Estimation}

\begin{figure}[h]
\begin{center}
\begin{tabular}{ccc}
    \begin{minipage}[b]{0.3\textwidth}
        \includegraphics[width=\textwidth]{images/Complexity/Rade-MNIST-FCNN-Epochs=4000-p=20-U.png}
    \end{minipage} &
    \begin{minipage}[b]{0.3\textwidth}
        \includegraphics[width=\textwidth]{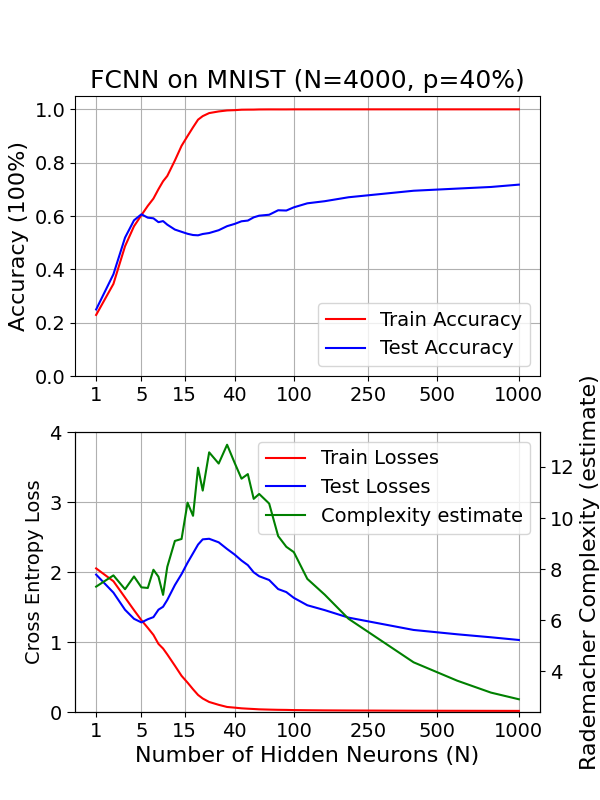}
    \end{minipage} &
    \begin{minipage}[b]{0.3\textwidth}
        \includegraphics[width=\textwidth]{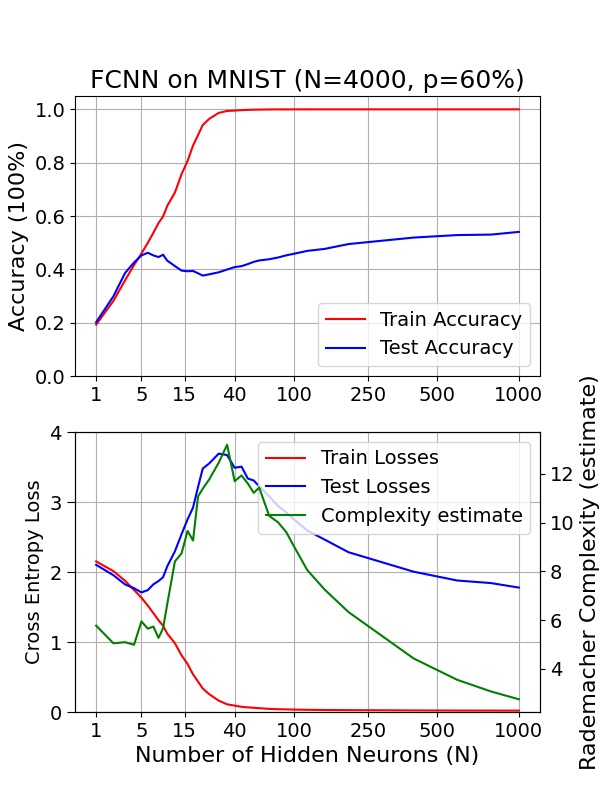}
    \end{minipage} \\
    \fontsize{8}{8}\selectfont (a) FCNN on MNIST $(p=20\%)$ &
    \fontsize{8}{8}\selectfont (b) FCNN on MNIST $(p=40\%)$ &
    \fontsize{8}{8}\selectfont (c) FCNN on MNIST $(p=60\%)$ \\
\end{tabular}
\end{center}
\caption{\label{fig:Rade-FCNN-MNIST-Noise}
    The mean estimated richness of class-wise activations of two-layer FCNNs trained on MNIST $(N=4000)$ across various widths with different label noise $p=[20\%, 40\%, 60\%]$. With a high label noise level, the trend in estimated complexity closely mirrors the test losses across all model widths $k$.
}
\end{figure}

Figure \ref{fig:k-NN-FCNN-MNIST-Noise} demonstrated the estimated complexities under high label noise ratio ($p = [40\%, 60\%]$). We may conclude that with a higher label noise ratio, the estimated complexity more precisely exhibits the double descent phenomenon, as the corresponding test loss curve. It is again noticeable that the numerical values of this estimated complexity dropped at the under-parameterized regime and increased in the peak value w.r.t. increase in noise. Nevertheless, the richness of class-wise activations of over-parameterized networks dropped significantly to near zero, which may pose that back-propagation through SGD might aim to minimize richness in over-parameterized networks. 

\begin{figure}[h]
\begin{center}
\begin{tabular}{ccc}
    \begin{minipage}[b]{0.3\textwidth}
        \includegraphics[width=\textwidth]{images/Complexity/Rade-CIFAR-10-CNN-Epochs=200-p=20-U.png}
    \end{minipage} &
    \begin{minipage}[b]{0.3\textwidth}
        \includegraphics[width=\textwidth]{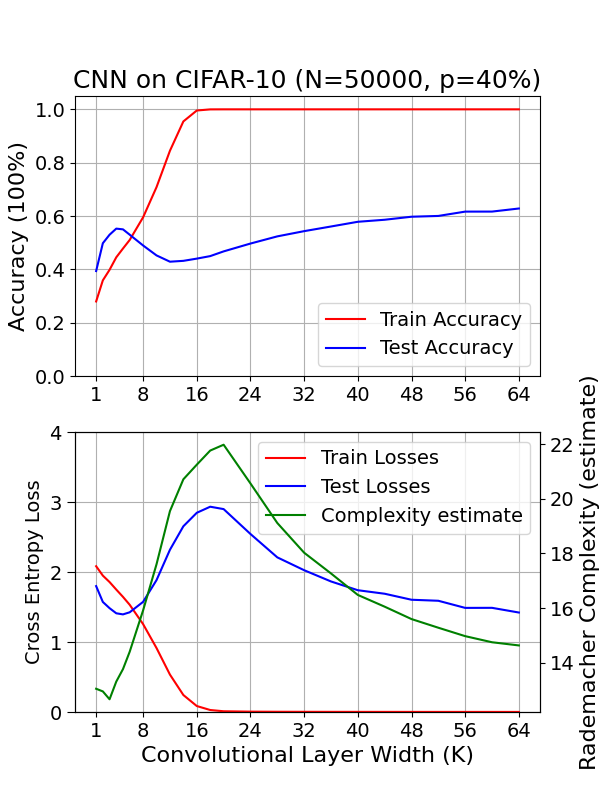}
    \end{minipage} &
    \begin{minipage}[b]{0.3\textwidth}
        \includegraphics[width=\textwidth]{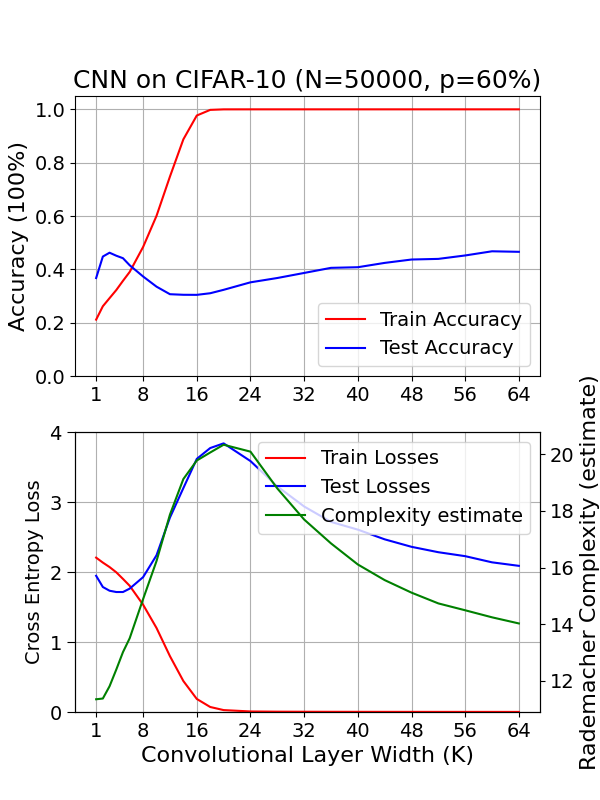}
    \end{minipage} \\
    \fontsize{8}{8}\selectfont (a) CNN on CIFAR-10 $(p=20\%)$ &
    \fontsize{8}{8}\selectfont (b) CNN on CIFAR-10 $(p=40\%)$ &
    \fontsize{8}{8}\selectfont (c) CNN on CIFAR-10 $(p=60\%)$ \\
\end{tabular}
\end{center}
\caption{\label{fig:Rade-CNN-CIFAR-10-Noise}
    The mean estimated richness of class-wise activations of five-layer CNNs trained on CIFAR-10 $(N=50000)$ across various widths with different label noise $p=[20\%, 40\%, 60\%]$. With a high label noise level, the trend in estimated complexity closely mirrors the test losses across all model widths $k$.
}
\end{figure}

The same conclusion can be derived from the experiment setup on five-layer CNNs, in which we can see an even stronger correspondence between the estimated class-wise activation richness (complexity) and the test error curve. While the interpolation threshold shifted rightwards as the label noise ratio increased, the complexity curve also shifted and dropped in numerical values as previous experiments showed. The differences existed in the under-parameterized regime that in the underfitting stage, with an increase in noise ratio, the drop in model complexity lasted shorter, resulting in an earlier increase in model complexity than the test error.

\section{Interpolation of Noise Data in the Feature Space}

\begin{figure}[h]
\begin{center}
\begin{tabular}{ccc}
    \begin{minipage}[b]{0.3\textwidth}
        \includegraphics[width=\textwidth]{images/Noise/k-NN-MNIST-FCNN-Epochs=4000-p=20-U.png}
    \end{minipage} &
    \begin{minipage}[b]{0.3\textwidth}
        \includegraphics[width=\textwidth]{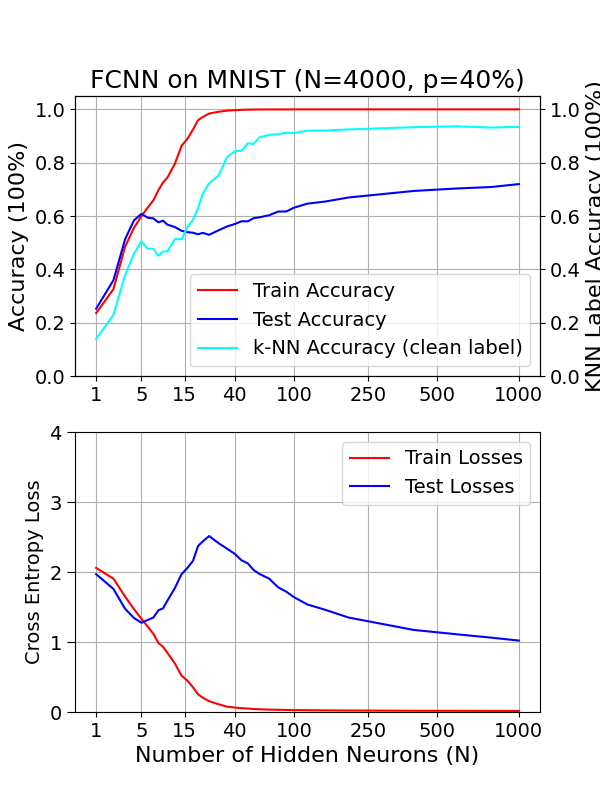}
    \end{minipage} &
    \begin{minipage}[b]{0.3\textwidth}
        \includegraphics[width=\textwidth]{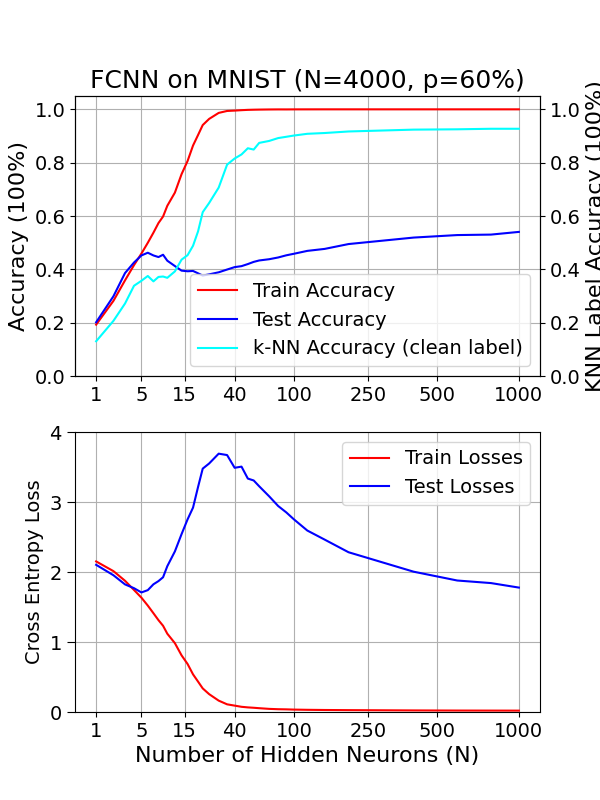}
    \end{minipage} \\
    \fontsize{8}{8}\selectfont (a) FCNN on MNIST $(p=20\%)$ &
    \fontsize{8}{8}\selectfont (b) FCNN on MNIST $(p=40\%)$ &
    \fontsize{8}{8}\selectfont (c) FCNN on MNIST $(p=60\%)$ \\
\end{tabular}
\end{center}
\caption{\label{fig:k-NN-FCNN-MNIST-Noise}
     The phenomenon of double descent on two-layer FCNNs trained on MNIST $(N=4000)$, under varying explicit label noise ratios of $p = [20\%, 40\%, 60\%]$ and the prediction accuracy of noisy labelled data denoted as $P$. 
}
\end{figure}

Figure \ref{fig:k-NN-FCNN-MNIST-Noise} presented the k-NN prediction test results on class-wise interpolation strategy on noisy data. We may see with a high label noise ratio, the drop in the overfitting stage turns to a small fluctuation, then continues increasing until passing the interpolation threshold and reaches 90\%. We believe our conclusion still applies that through correctly interpolating noisy samples among correct samples, over-parameterized networks gain the ability to avoid over-fitting on noisy information. However, as label noise increases, effectively isolating noise becomes more challenging. Correctly classifying between true samples does not necessarily indicate superior testing performance. This is because the reduced number of true samples results in sparser representations in the hidden feature space, leading to unreliable prediction results. Again, the objective of this test is to offer insights into why over-parameterized networks outperform under-parameterized counterparts when fully fitting the training data with extensive noise information, rather than serving as a predictor of generalization performance.

\begin{figure}[h]
\begin{center}
\begin{tabular}{ccc}
    \begin{minipage}[b]{0.3\textwidth}
        \includegraphics[width=\textwidth]{images/Noise/k-NN-CIFAR-10-CNN-Epochs=200-p=20-U.png}
    \end{minipage} &
    \begin{minipage}[b]{0.3\textwidth}
        \includegraphics[width=\textwidth]{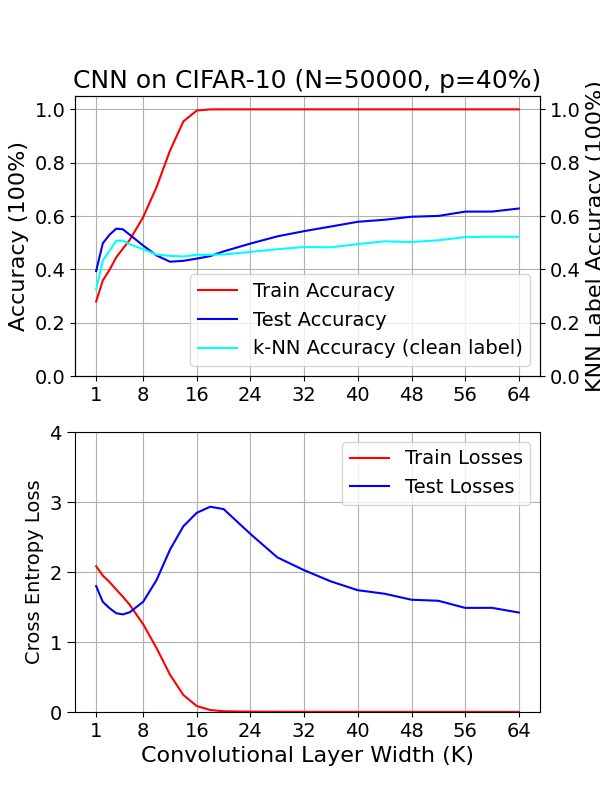}
    \end{minipage} &
    \begin{minipage}[b]{0.3\textwidth}
        \includegraphics[width=\textwidth]{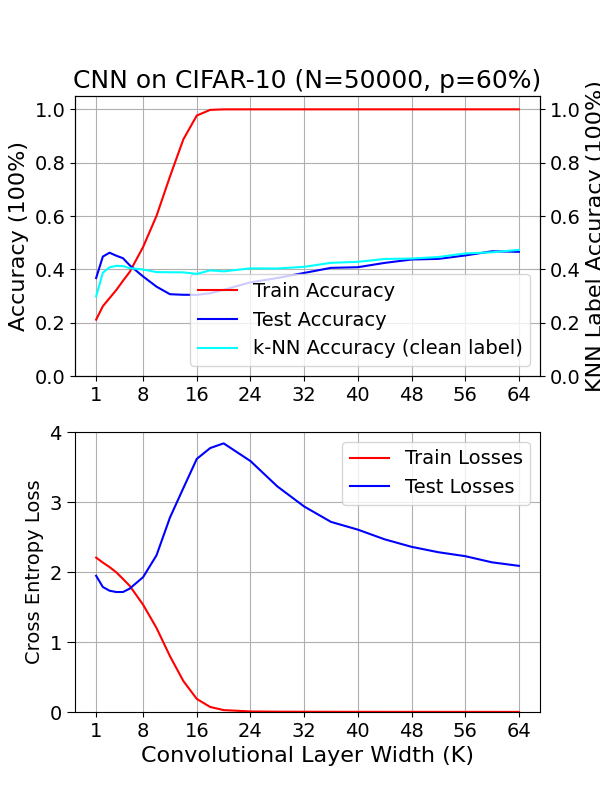}
    \end{minipage} \\
    \fontsize{8}{8}\selectfont (a) FCNN on MNIST $(p=20\%)$ &
    \fontsize{8}{8}\selectfont (b) FCNN on MNIST $(p=40\%)$ &
    \fontsize{8}{8}\selectfont (c) FCNN on MNIST $(p=60\%)$ \\
\end{tabular}
\end{center}
\caption{\label{fig:k-NN-CNN-CIFAR-10-Noise}
     The phenomenon of double descent on five-layer CNNs trained on CIFAR-10 $(N=50000)$, under varying explicit label noise ratios of $p = [20\%, 40\%, 60\%]$ and the prediction accuracy of noisy labelled data denoted as $P$. 
}
\end{figure}

With an increased label noise ratio introduced on CIFAR-10 and trained with five-layer CNNs, we may observe that the overall change in prediction accuracy $P$ is less significant and even neglectable with $p=60\%$. We attribute this observation to the reason that more than half of the training samples are considered noisy and assigned random labels, the sparsity between the clean samples in any feature space could be expected, resulting in loose constraints on noisy samples. Thus the reduction in interpretability of k-NN predictions could be expected. While for the test case with $p=60\%$, the k-NN prediction accuracy `coincides' with the test accuracy, we attribute that there is no connection between the two values. We would like to re-emphasize the discussion we made in Chapter \ref{chapter4:complexity} that the hypothesis behind this experiment assumes the learned representations more of an injection between feature spaces. When multiple injections are performed and more abstract features are learned akin to their assigned class labels, the hypothesis would fail (like the experiments on ResNet18s). Nevertheless, we expect the experiment results could provide insights into the interplay of neural networks fitting noisy information.

\section{Implicit Activation Sparsity}

\begin{figure}[h]
\begin{center}
\begin{tabular}{ccc}
    \begin{minipage}[b]{0.3\textwidth}
        \includegraphics[width=\textwidth]{images/Sparsity/Act-Ratio-MNIST-FCNN-Epochs=4000-p=20.png}
    \end{minipage} &
    \begin{minipage}[b]{0.3\textwidth}
        \includegraphics[width=\textwidth]{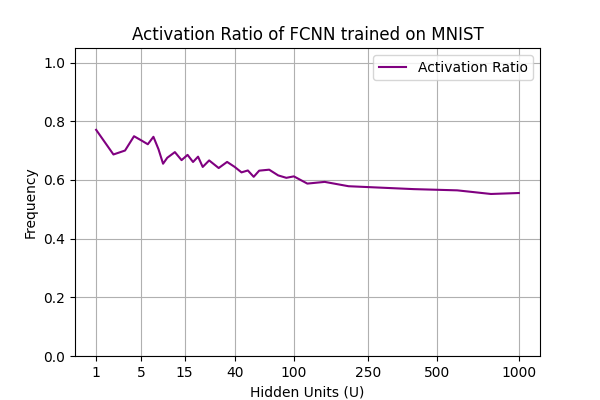}
    \end{minipage} &
    \begin{minipage}[b]{0.3\textwidth}
        \includegraphics[width=\textwidth]{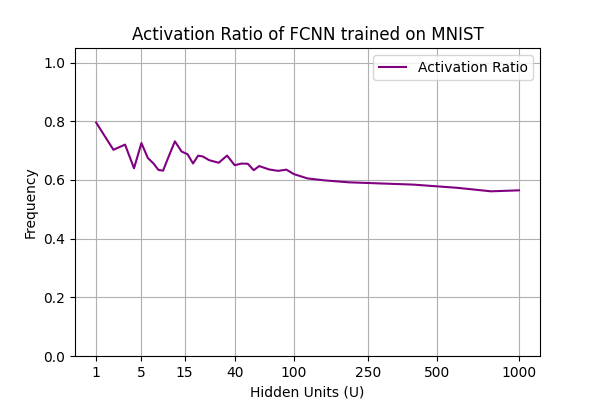}
    \end{minipage} \\
    \begin{minipage}[b]{0.3\textwidth}
        \includegraphics[width=\textwidth]{images/Sparsity/NDCG-MNIST-FCNN-Epochs=4000-p=20.png}
    \end{minipage} &
    \begin{minipage}[b]{0.3\textwidth}
        \includegraphics[width=\textwidth]{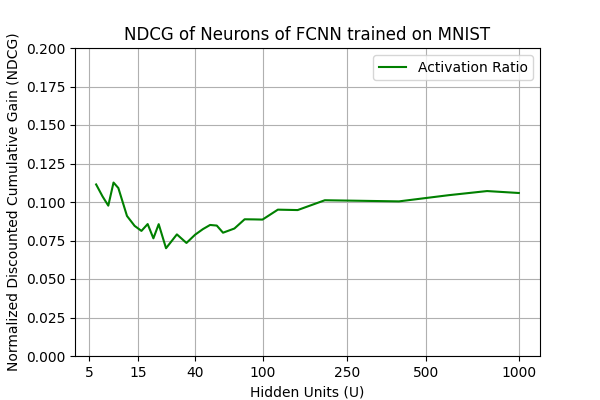}
    \end{minipage} &
    \begin{minipage}[b]{0.3\textwidth}
        \includegraphics[width=\textwidth]{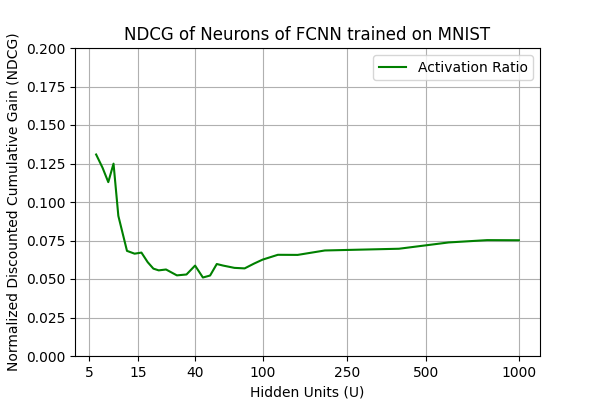}
    \end{minipage} \\
    \fontsize{8}{8}\selectfont (a) FCNN on MNIST $(p=20\%)$ &
    \fontsize{8}{8}\selectfont (b) FCNN on MNIST $(p=40\%)$ &
    \fontsize{8}{8}\selectfont (c) FCNN on MNIST $(p=60\%)$ \\
\end{tabular}
\end{center}
\caption{\label{fig:Sparsity-FCNN-MNIST-Noise}
    The mean activation ratio of hidden neurons and the class-wise NDCG of two-layer FCNNs trained on MNIST, under varying explicit label noise ratios of $p = [20\%, 40\%, 60\%]$. The fluctuation in activation sparsity was less pronounced with increased label noise, while the class-wise NDCG demonstrates a `U-shaped' curve, decreasing before the interpolation threshold and increasing thereafter.
}
\end{figure}

Since activation sparsity only presented in FCNNs than CNNs, we still focus on discussing the experiment results with two-layer FCNNs trained on MNIST. Figure \ref{fig:Sparsity-FCNN-MNIST-Noise} presented the experiment on the activation sparsity of neurons, with an increased label noise ratio ($p = [40\%, 60\%]$). The figure suggests that while the fluctuation in activation sparsity remains within a certain range, it becomes more concentrated with increased noise. This could be attributed to the increased difficulty in minimizing activation complexities. Conversely, the experimental results of neural NDCG on class prediction frequency are intriguing: in the under-parameterized regime, neural class specialization decreases exponentially, only to rise again after the interpolation threshold. The decrease in neural NDCG values may be associated with the increase in class-wise activation correlation between the representation layer depicted in Figure \ref{fig:FCNN-MNIST-Noise}. In conclusion, we may again emphasize that activation sparsity is not likely to be a main cause of the double descent phenomenon, while the neural NDCG might provide valuable insights into neural network's `preferences' on how to memorize feature patterns with more `specialized' neurons in the over-parameterized regime.

\chapter{Exploring Depth vs. Width: Neural Network Architectural Choices\label{appendix:architecture}}
In this study, we have focused on scaling parameterization on model width with the same depth for each neural architecture. An easy concern is that this cannot account for discussions over different architectural choices, especially including the comparison between a deeper or wider one. Nevertheless, owing to the powerful modelling capabilities of neural networks and the relatively straightforward nature of image classification tasks on MNIST, employing 20 hidden neurons enables complete interpolation on the training dataset with an input dimension of 784. With standard deep learning setups, it is hard to compare neural networks with different layer widths on the same scaling of the number of parameters. 

Thus, a special experiment is conducted with a prefixed representation layer of 20 hidden neurons as the output and random weights set on MNIST. Fully connected classifier layers of depth 1-4 are connected to this representation layer, to learn and perform feature extraction from the `random' representations of MNIST training images. In multi-layer classifiers, the number of hidden neurons on each layer remains consistent. During the training stage, only the weights of the classifier layers are trained and updated for 400 epochs, while the representation layer remains unchanged. Under this experiment setup, we can compare different architectural choices under the same scales of the number of parameters. 

\begin{figure}[ht]
    \centering
    \includegraphics[width=0.6\linewidth]{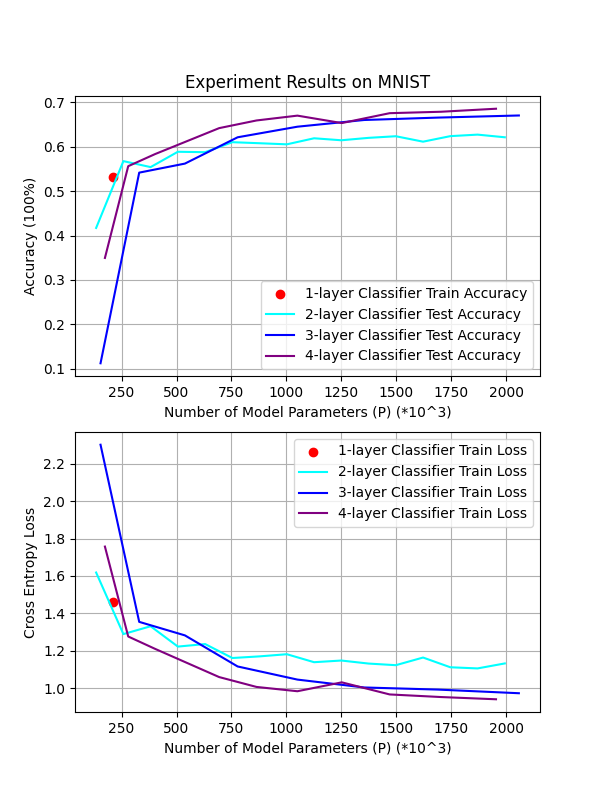}
    \caption{Experiment results with a prefixed untrained random representation layer and classifier of 1-4 layers trained on MNIST. All layers are fully connected with no regularization technique applied and trained for a consistent 400 epochs. The results demonstrated that with a sufficient number of parameters, deeper networks are better than wider ones. }
    \label{fig:seperate-layer}
\end{figure}

The experiment results are presented in Figure \ref{fig:seperate-layer} from which we may conclude that with guaranteed sufficient neural dimension to perform effective information extraction, deeper networks are better than wider ones under the same number of model parameters. Deeper networks perform poorer than wider networks or even with one-layer classifiers can be caused by insufficient neural dimension for information representation, in particular the 10\% accuracy with 4 hidden neurons on each layer in a four-layer neural network. While in the main context of the paper we have demonstrated the effectiveness of over-parameterization with rich dimensions, the take-away message from this experiment is that adding parameters and model depth could lead to potential best practices.

\end{document}